\pdfoutput=1
\documentclass[11pt]{article}

\usepackage[]{ACL2023}

\usepackage{times}
\usepackage{latexsym}

\usepackage[T1]{fontenc}
\usepackage[utf8]{inputenc}

\usepackage{microtype}

\usepackage{inconsolata}

\usepackage{times}
\usepackage{latexsym}
\usepackage{subcaption}
\usepackage{hhline}
\usepackage{booktabs}
\usepackage{multirow}
\usepackage{amsmath}
\usepackage{graphicx}
\usepackage{color}
\usepackage{makecell}
\usepackage{diagbox}
\usepackage{enumerate}
\usepackage{amssymb}

\title{Reducing Sensitivity on Speaker Names for Text Generation from Dialogues}
\author{Qi Jia$^1$, Haifeng Tang$^2$, Kenny Q. Zhu$^3$\thanks{\hspace{2mm}The corresponding author.}\\
	$^{1,3}$Shanghai Jiao Tong University, Shanghai, China \\
	$^2$China Merchants Bank Credit Card Center, Shanghai, China \\
	\texttt{$^1$Jia\_qi@sjtu.edu.cn}, 
	\texttt{$^2$thfeng@cmbchina.com}, 
	\texttt{$^3$kzhu@cs.sjtu.edu.cn}\\
}

\begin{document}
\maketitle
\begin{abstract}
%The semantics of most dialogues are not grounded on the speakers names,
%therefore 
%close-form dialogue generation. 
Changing speaker names consistently throughout a dialogue should not 
affect its meaning and corresponding outputs for text generation from dialogues. 
However, pre-trained language models, serving as
the backbone for dialogue-processing tasks, have shown to be sensitive to 
nuances. This may result in unfairness
in real-world applications. 
No comprehensive analysis of this problem has been done in the past.
In this work, we propose to quantitatively measure a model's sensitivity 
on speaker names, and comprehensively evaluate a number of known methods for reducing speaker name sensitivity,
including a novel approach of our own. 
Extensive experiments on multiple datasets provide a benchmark for 
this problem and show the favorable performance of our approach in 
sensitivity reduction and quality of generation.  
\end{abstract}

\section{Introduction}
\label{sec:intro}
% dialogue -> fairness in dialogues => dialogue understanding tasks
%Dialogues as the most natural way for information exchanges has gained great attention and its research directions can be divided into two categories. One is open-ended dialogue systems~\cite{gu2021dialogbert,xu2021learning} which aims at generating or selecting appropriate responses to fulfill the users' needs. The other is dialogue understanding tasks which helps to quickly digest information and returns the expected output given the whole dialogue. 

%With the prosperous of fine-tuning with pre-trained language models, dramatic improvements have achieved on a range of dialogue tasks, pushing their implementation in practice. 
%\JQ{response generation not applicable; static dialogue generation}
%\JQ{sensitivity of speaker names ia a part of fairness issue ?}
The safety and fairness issue of generations from 
dialogue models is a crucial concern in real applications. 
Previous work focuses on response generation from open-ended dialogue systems~\cite{xu2020recipes, henderson2018ethical}, such as offensive contents~\cite{baheti2021just}, gender bias~\cite{liu2020does,dinan2020queens} and other discriminated behavior~\cite{sheng2021revealing,smith2021hi}. For other text generation tasks where the whole dialogue is provided and the 
output shouldn't go beyond the dialogue, such as dialogue summarization~\cite{gliwa2019samsum} and dialogue reading comprehension~\cite{li2020molweni},
the fairness issue is still unexplored. %\JQ{more example tasks}

% leaving the fairness of close-style dialogue understanding tasks unexplored.

\begin{figure}[th]
	\centering
	\includegraphics[width=0.9\columnwidth]{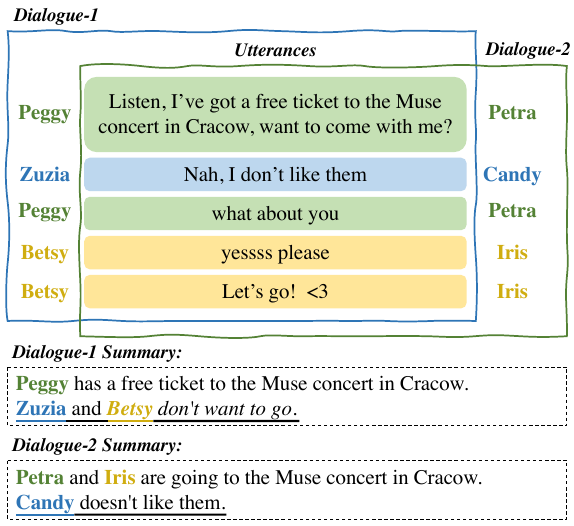}
	\caption{Two instances of an example from the SAMSum dataset, each
with a different set of names. Two different summaries are generated by BART. 
Different colors indicate different speakers. \underline{divergent contents} are underlined and \textit{incorrect contents} are italicized.} %in the summaries.}
%\underline{Divergent contents} are underlined in the generated summaries.}
	\label{fig:example}
\end{figure}

% dialogue understanding tasks, characteristics  with example
%Representative dialogue understanding tasks includes dialogue summarization~\cite{lin2022other,zhong2022dialoglm}, reading comprehension~\cite{sun2019dream,li2020molweni}, and etc. 
In these tasks, the input dialogues are self-contained, and the names of the speakers
do not carry any connotation from outside of the dialogue. 
Therefore,
changing the speaker names consistently in a dialogue should not affect the 
meanings of the dialogue and the desired outputs.
This contrasts with response generation, where the dialogue is in progress and the output is expected to be different in styles or contents for various speakers.
Taking dialogue summarization~\cite{gliwa2019samsum,chen2021dialogsum} as an example for text generation from dialogues, it focuses on
%An example for text generation from dialogues is dialogue summarization tasks in \citet{gliwa2019samsum} and \citet{chen2021dialogsum}, focusing on
generating concise ``who-did-what'' summaries in the third person.
In Fig.~\ref{fig:example},
%Taking the dialogues in 
%Fig.~\ref{fig:example} as an example, 
the two dialogues are identical except for the speaker names. 
The two summaries are expected to be the same modulo 
the speaker names. 
%In conclusion, speaker name insensitivity is an inherent and obvious characteristic of dialogue understanding.

% robustness of language model
Unfortunately, models nowadays, following the pretrain-finetune paradigm, 
are sensitive to trivial changes, which has been verified in other tasks. 
In relation extraction, spurious correlations between entity mentions and 
relations lead to entity bias~\cite{zhang2018graph,zhang-etal-2017-position,wang-etal-2022-rely}. 
%Some work~\cite{zhang2018graph,zhang-etal-2017-position} proposes to 
%prevent it by masking entities during fine-tuning. 
%\citet{wang-etal-2022-rely} debiases the models by removing the 
%counterfactual predictions based on causal inference. 
Other similar work includes the analysis of robustness by entity 
renaming for machine reading comprehension models on narrative 
texts~\cite{yan2022robustness} and name biases in machine translation 
with inflected languages~\cite{wang2022measuring}, like German. 
Besides, \citet{shwartz2020you} claims that pre-trained language models do not treat given names as interchangeable or anonymous, showing unfairness in reading comprehension.% by switching names with pre-defined templates. % with span-based or classification models 
%\JQ{add ~\cite{shwartz2020you} Pre-trained LMs do not treat given names as interchangeable or anonymous: This has not only implications for the quality and accuracy of systems that employ these LMs, but also for the fairness of those systems.}

% sensitivity of PrLMs in dialogue understanding tasks
Obviously, dialogue understanding models are sensitive to speaker names 
according to Fig.~\ref{fig:example} as well. 
The model tends to generate different information 
given different speaker names, such as ``don't want to go'' and 
``doesn't like them''.
Incorrect content, 
``... Betsy don't want to go'', is generated with the first group of speakers, 
while not with the other group. 
%``Ashley is on her way'' and 
%``Derek will bring the dog''.  
According to our pilot experiment with the vanilla BART fine-tuned on SAMSum, around 74.00\% of generations are changed by switching speaker names and 69.82\% among them are due to distinct contents.
Such uneven performances %can lead to differences on information allocation and 
create unfairness among 
different speakers, especially in the aspect of information allocation. 
The model may also catch latent properties 
in names~\cite{romanov2019s} and lead to discrimination, %against specific groups, 
raising the importance of research on the sensitivity on speaker names.
%Different from named entities, 
%\KZ{Don't understand this: 
%speakers in these tasks are more isolated 
%from the dialogue contents and have closer connections to users in real applications}, 

Previous work has also mentioned this problem. Different data pre-processing approaches are adopted during the construction of datasets to avoid using speaker names, such as ``A'' or ``B'' in \citet{li2017dailydialog}. \citet{khalifa2021bag} replace speaker names with more common and frequent names that the model may have seen during pre-training. Data augmentation by changing speaker names is adopted by \citet{liu2021controllable}.
However, all of them only attempted to attack this problem 
subjectively, without quantitive analysis and fair comparisons. %these works

% approach classification / our approach
In this work, we systematically analyze speaker name sensitivity in text generation from dialogues. We define the speaker name sensitivity and divide the approaches 
into offline and online ones. 
Then, we propose two novel insensitivity losses, helping to reduce attention and hidden state distances of the same dialogue with different speaker names for transformer-based models during fine-tuning. These losses can be used in both kinds of approaches.
Results on several tasks show that
%online approach which replaces names with frequent ones achieves the best generation performance, and coupled with data augmentation, it provides
%the best fairness among all baselines.
our losses reduce the sensitivity and get better generations. %on task-specific metrics. %, achieving the state-of-the-art performance.
%\KZ{I'm a little confused reading the above two sentences. Which one is better,
%the freq or our ins loss?}
% contributions
In summary, our contributions are:
\begin{itemize}
	\item We are the first to investigate the speaker name sensitivity 
in text generation from dialogues (Sec.~\ref{sec:problem}) with all of the codes and results open-sourced at \url{https://github.com/JiaQiSJTU/SpeakerNameSensitivity}.% and quantify the performance sensitivity for further analysis
	\item We introduce two novel insensitivity losses as auxiliary training objectives for reducing sensitivity during fine-tuning (Sec.~\ref{sec:approach}).
	\item Experiments on different tasks provide a benchmark with comprehensive analysis on speaker name sensitivity, and show  state-of-the-art performances of our approach
%stronger overall performance with lower speaker name sensitivity 
(Sec.~\ref{sec:results}).
\end{itemize}

\section{Background}

\subsection{Speaker Name Sensitivity}
\label{sec:problem}

\textit{Speaker name sensitivity} is the differences in the generations by a model, given the 
identical dialogues except for different speaker names. We define it as follows.
%We setup the problem specifically as follows. %the same dialogue utterances

%We first define the speaker name sensitivity of a model on a sample as follows:

Let $d$ denote the input dialogue. $c$ denotes other input content, which can be empty for tasks like dialogue summarization, or a piece of text such as a question for reading comprehension. $p$ refers to the set of speakers names in $d$. $f$ is a one-to-one mapping which maps $p$ into a set of names $p'$ from a name pool $\mathcal{P}$ consisting of a set of candidate names to be substituted into the samples. The names $p'$  are sampled under the uniform distribution without the loss of generality.
\textit{The speaker name sensitivity $SS$ of a generation model $\mathcal{M}(\cdot)$ on this sample} is:
\begin{equation}
	\label{eq:ss}
	\begin{aligned}
	SS(\mathcal{M} | d, c) = \delta(&\{\mathcal{M}(Rep(d, c | f )) \\
	&| \forall f: p\rightarrow p', p'\subseteq \mathcal{P} \})
	\end{aligned}
\end{equation}
where $Rep(\cdot)$ replaces names in the sample given $f$, i.e., from $p$ to $p'$. $\delta(\cdot)$ quantifies the differences among generations.

Then, \textit{the sensitivity $SS$ of a model $\mathcal{M}(\cdot)$} is the expectation $\mathbb{E}$ of over all samples from the real-world distribution $D$:
 \begin{equation}
 		SS(\mathcal{M}) = \mathbb{E}_{(d,c)\sim D} [ SS (\mathcal{M} | d,c) ] 
 \end{equation}

In practice, a dialogue dataset is regarded as a sampling from $D$ for evaluations.  Each sample in the dataset is provided with a reference output $o$ for supervised training. We use $D_{tr}$, $D_{va}$ and $D_{te}$ to refer to training, validation and test sets.
See detailed implementations and metrics in Sec.~\ref{sec:quatification}.

\subsection{Existing Approaches}

We investigate existing approaches that target on reducing the sensitivity and classify them into offline ones and online ones, where the former chases to reduce the sensitivity by exploring better model parameters and the latter pursues insensitivity by unification or simplification of input data. Thus, data processing steps are required before inputting into the model and after the inference during the test time and speaker names in $D_{tr}$, $D_{va}$ and $D_{te}$ are all changed for online approaches.
The model needs fine-tuning for both approaches. 

%Offline approaches fine-tune models to be more insensitive to speaker names. 
%No extra steps are required when using the model during the test time.

\textit{Offline approaches} include:

\textbf{Embedding Layer(Emb)}: Similar to~\cite{gu2020speaker} and ~\cite{heetal2021speakerturn}, an additional embedding layer can be adopted for representing whether the model should be sensitive to corresponding tokens. $2$ embeddings are learned during fine-tuning. 

\textbf{Augmentation (Aug)}: \citet{liu2021controllable} proposed to do data augmentation by exchanging speaker names in training samples with names from $D_{tr}$. They aim to reduce unexpected inductive bias caused by speaker names, which is similar to our goal. The model is fine-tuned with augmented training data while $D_{va}$ and $D_{te}$ remain unchanged.

%\textbf{Insensitivity Loss (Ins)}: We newly propose a fine-tuning objective on top of the augmented data in Fig.~\ref{fig:insloss}. More details are in Sec.~\ref{sec:insloss}.% shown in Figure~\ref{fig:insloss}, helping to reduce sensitivity with augmented data (More in ~\ref{sec:insloss}).

%\subsection{Online Approaches}
%\label{sec:dda}
\textit{Online approaches} are:
%also need fine-tuning. The main difference is that data processing steps are required before inputting into the model and after the inference during the test time. Speaker names in $D_{tr}$, $D_{va}$ and $D_{te}$ are all changed.%Two approaches are as follows.

\textbf{ID:} Some works~\cite{cui2020mutual,li2017dailydialog} replace speaker names with predefined IDs to avoid name bias.
We use ``Speaker[NUM]'' similarly to \citet{kim2019eighth} and \citet{chen2021dialogsum}, 
which is close to words seen during pre-training and fits different numbers of speakers.
``[NUM]'' is the index of a speaker's first occurrence. 

\textbf{Frequent (Fre)}: This refers to the approach proposed in \citet{khalifa2021bag}. They use 100 frequent male and 100 frequent female names online\footnote{https://www.ssa.gov/oact/babynames/decades/century.html} as the pool $P$ for sampling replacements. This approach can be combined with Aug into \textbf{FreAug}.

\section{Proposed Approach}
\label{sec:approach}

%\begin{figure*}[th]
%	\centering
%	\includegraphics[scale=0.6]{insloss.pdf}
%	\caption{An illustration of our Insensitivity Loss when $K=2$.
%	}
%	\label{fig:insloss}
%\end{figure*}

%Previous approach Aug only tries to reduce undesired biases with more data while neglects the model itself. 
We focus on the widely-accepted encoder-decoder architecture for pre-trained generation models and design two auxiliary insensitivity losses to take full advantage of augmented data on top of Aug. 
Given the dialogue sample with different speaker names, a model outputs distinct generations due to its different internal behaviors. Therefore, penalizing unexpected internal differences should help the model behave consistently and reduce the sensitivity.
 
With this intuition, we propose the cross-attention loss and the decoder-hidden-state loss. An illustration for them is in Appendix~\ref{sec:app-model}. 
The former corresponds to cross-attention distributions that help the decoder make a soft information selection among encoder hidden states at each step and should be similar with different speaker names. 
%Thus, narrowing down the differences of such hidden information selection reflected by attention distributions is necessary. 
The latter is based on the final decoder hidden states which are expected to be the same under the default teacher-forcing training strategy except for the speaker name tokens.
We didn't consider the encoder attentions since according to our pilot analysis of the vanilla BART, %with pairs of samples, 
the cross attentions distance of the different predictions is around 1.5 times of the same ones. However, there are no differences in the encoder attentions. 
Other intermediate hidden states are excluded since they are all affected by different input embeddings of speaker names, except that the final decoder hidden states are sure to be the same.% to get identical predictions. %More details are as follows. %detailed explanations for two losses are as follows.

%The main intuition is to penalize the unexpected differences of model's internal behavior among the same dialogue samples with different speaker names.

%On top of the previous approach Aug, we propose two new auxiliary losses to train more insensitive models: the cross-attention insensitivity loss and the decoder-hidden-state insensitivity loss. The intuition for both losses are to penalize the differences of model's internal behavior among the same dialogue samples with different speaker names.
%More details are as follows.
%Then, we propose two insensitivity losses to further reduce the sensitivity. 

%The encoder-decoder architecture is the most widely-accepted design for pre-trained generation models. 
%At each time step, the decoder selects information to concentrate on among the encoder hidden states with the cross-attention mechanism, and predicts the current output. This soft information selection process determines the contents included in the final predictions. 
%The intuition here is to narrow down the differences of such hidden information selection reflected by attention distributions.

\subsection{Cross-attention Insensitivity Loss}
\label{sec:insloss}

%\JQ{design space: input, cross attention, decoder}
%We did a pilot experiment on the SAMSum dataset with vanilla fine-tuned BART. By switching names, we collect two generations for each sample. Among the 71.92\% samples with different outputs, we randomly sampled 200 pairs and asked human annotators to label each case among 4 types (More details are in the Appendix). The Kappa score is 95.94\% with perfect agreement. 

%According to the pilot experiment metioned in Sec.~\ref{sec:intro}(More details are in the Appendix), the results shows that the most severe sensitivity is from content distinction, meaning that generations contain different information or keywords. This inspired us to design the following loss function.

%We denote the input and output of a transformer-based model as $X=\{x_1, x_2, ..., x_{din} \}$ and $Y=\{y_1, y_2, ..., y_{dout}\}$ respectively. $din$ and $dout$ represent the number of corresponding tokens.
We denote a model's input and output length, i.e., the number of tokens, as $din$ and $dout$.
During training, the cross attentions calculated for each output token are collected as $CA\in R^{N\times dout \times din}$. $N$ is the number of heads for the multi-head attention mechanism, determined by the configuration of pre-trained models. We apply average pooling over the dimension of $dout$, to get the overall attention over the input tokens $\overline{CA}\in R^{N\times din}$.
% $\rm{avg}(\cdot)$

Given an original sample $\{d_i, c_i, o_i\}$, we construct $K-1$ augmented samples by replacing speaker names. The averaged attentions for all samples are $\{\overline{CA_k}\}_{k=1}^K$. Since it is a default that each sample should go through the tokenizer before inputting to the model, $\{din_k\}_{k=1}^K$ are not guaranteed to be identical in two cases.
%. The lengths of the two distributions are different in two cases. 
First, names may be tokenized into different token counts. For example, ``John'' and ``Robinson'' are tokenized into \{``John''\} and \{``Rob'', ``inson''\} by BART tokenizer. Replacing ``John'' with ``Robinson'' in $d_i$ will increase the sequence length. Second, long inputs may be truncated at different tokens. So, we consider two corresponding functions for unification: %before calculating the distance of these distributions. %We consider two corresponding functions:
\begin{itemize}
\item $\rm{Sum(\cdot)}$ sums up the attention values of tokens belonging to an occurrence of a speaker name. 
\item $\rm{Pad(\cdot)}$ pads attentions into the same length $din_u$ by concatenating zeros, which means that this part of contents is missing.% due to truncation.
%\begin{itemize}
%	\item 
%	\item 
\end{itemize}
The unified $\{\overline{CA_k}\}_{k=1}^K$ is represented as $\{\widetilde{CA}_k\}_{k=1}^K$, where $\widetilde{CA}_k\in R^{N\times din_u}$.

Finally, the loss is calculated as:
\begin{equation}
	\mathcal{L}_{ca} = \frac{1}{K(K-1)} \sum_{k=1}^{K}\sum_{l=1, l\neq k}^{K} {loss}(\widetilde{CA}_k, \widetilde{CA}_l)
\end{equation}
where ${loss}(\cdot)$ measures the distances between a pair of attentions. 

\subsection{Decoder-hidden-state Insensitivity Loss}

Similarly, hidden states of the decoder's final output for all samples can be denoted as $\{DH_k\}_{k=1}^K$, where $DH_k\in R^{dout_k\times H}$ and $H$ represents the hidden size. The lengths of them also vary due to the above two cases. We adopt two different functions: %for unification:
\begin{itemize}
	\item $\rm{Del(\cdot)}$ ignores the hidden states whose predicted tokens belong to a speaker name.
	\item $\rm{Trunc(\cdot)}$ truncates the redundant hidden states at the end without the paired ones.
\end{itemize}
Thus, the unified $\{DH_k\}_{k=1}^K$ is represented as $\{\widetilde{DH}_k\}_{k=1}^K$, where $\widetilde{DH}_k\in R^{dout_u\times H}$.

The loss is defined as:
\begin{equation}
	\mathcal{L}_{dh} = \frac{1}{K(K-1)} \sum_{k=1}^{K}\sum_{l=1, l\neq k}^{K} loss(\widetilde{DH}_k, \widetilde{DH}_l)
\end{equation}
We adopted the mean square error for both losses.

\subsection{Learning Objective}

$\mathcal{L}_{ca}$ and $\mathcal{L}_{dh}$ are added to the vanilla generation loss $\mathcal{L}_{gen}$ with hyper-parameters $\alpha$ and $\beta$: %newly-introduced
\begin{equation}
	\begin{aligned}
		%\mathcal{L}_{gen} =  \frac{1}{K}\sum_{k=1}^{K} \rm{loss}_{gen}(d_i^k, o_i^k) \\
		\mathcal{L}_{total} = \mathcal{L}_{gen} + \alpha \mathcal{L}_{ca} + \beta \mathcal{L}_{dh}
	\end{aligned}
\end{equation}
The insensitivity losses are only auxiliary fine-tuning objectives, leaving the inference time unchanged. They can be added on top of both Aug and FreAug, denoted as \textbf{Ins} and \textbf{FreIns}.

%We also compared the attention distance in this way of output pairs mentioned above. It shows that the attention distance for pairs with different outputs is 1.3~2.3 times of the same ones.

\section{Experimental Setup}
% introduce the datasets, evaluation metrics of different tasks
We define the evaluation metrics for sensitivity, introduce multiple text generation tasks with dialogue data and present implementation details.%at the end

\subsection{Evaluation Metrics for Sensitivity}
\label{sec:quatification}

We uniformly sample names from $P$, which is specified later, to realize $f$ without the loss of generality and re-sample the name if it is not in $p$ but in the conversation. We avoid changing names mentioned during the conversation in case they are grounded entities. Since it's impossible to enumerate all possible $f$, we choose to substitute names of samples in $D_{te}$ for $T=5$ times. It should be noted that varying names in test data is different from the augmentation approach. The additional test data is fixed once constructed for comparing approaches by quantitatively measuring the sensitivity.

%under the uniform distribution
%The divergence of the predicted results among $D^t_{te}|_{t=1}^T$ for each sample reflects the model's sensitivity to speaker names.
%In a word, the speaker name sensitivity of a model can be reflected by the differences in generations or the variation of scores averaged with a number of samples from $D_{te}$(see metrics in Sec.~\ref{sec:quatification}). 

%By replacing names in $D_{te}$ for $T$ times, we can construct multiple test sets. The divergence of the predicted results among $D^t_{te}|_{t=1}^T$ for each sample reflects the model's sensitivity to speaker names. It should be noted that varying names in test data is different from the augmentation approach. The additional test data is fixed once constructed and is used to compare different approaches by quantitatively measuring the sensitivity.
%Dialogue models are expected to perform identically on different replaced samples and get the same scores with task-specific evaluation metrics compared with the reference $o$. 
%i.e., adding $o$ as the input of $Diff(\cdot)$ in Eq.~\ref{eq:ss}. 

%We define specific metrics for evaluating speaker name sensitivity similar to \citet{prabhakaran2019perturbation}' work.
We introduce three kinds of $\delta(\cdot)$ with task-specific evaluation metric $\rm{Score}(\cdot)$, such as Rouge and BertScore for dialogue summarization, and measure the speaker name sensitivity of a model similar to \citet{prabhakaran2019perturbation}' work.
\textbf{Pairwise Sensitivity(S-*)} is defined as:
\begin{equation}
	E_{i=1}^{N^{te}} E_{t_1=1}^{T}E_{t_2=1, t_1\neq t_2}^T[1-{\rm Score} (\hat{o}^{t_1}_i, \hat{o}^{t_2}_i)]
\end{equation}
$\hat{o}^t_i$ is the generation where replaced names are changed back for evaluation. $N^{te}$ is the number of samples in $D_{te}$. $E(\cdot)$ is the mean operator. 

Dialogue models are also expected to get the same scores with task-specific evaluation metrics compared with the reference $o$. So, we can also add $o$ as the input of $\delta(\cdot)$ in Eq.~\ref{eq:ss} and define the following two metrics:
\textbf{Score Range (R-*)} as
\begin{equation}
	\begin{aligned}
		E_{i=1}^{N^{te}} [& \max(\{{\rm Score}(o_i, \hat{o}^t_i)|_{t=1}^T\}) \\
		&-  \min(\{{\rm Score}(o_i, \hat{o}^t_i)|_{t=1}^T\})]
	\end{aligned}
\end{equation}
and \textbf{Score Deviation (D-*)} as
\begin{equation}
	E_{i=1}^{N^{te}} [ {\rm StdDev} (\{{\rm Score}(o_i, \hat{o}^t_i)|_{t=1}^T\}) ]
\end{equation}
%$o_i$ represents the reference and 
The sensitivity metrics here are the lower the better and are denoted by $\downarrow$ in the following sections. %can be combined with different task-specific metrics, and 

%If test data are constructed by switching all speaker names in each sample, we call it \textbf{change-all-name} test. We also only change the name of a single speaker each time to do \textbf{change-one-name} tests for analyzing fine-grained sensitivity.% and can be evaluated in the same way.

%change-all-name change-one-name

\subsection{Tasks and Datasets}

We implement our experiments on the tasks below. The statistics are in Table~\ref{tab:taskdata} and we calculate the macro-average scores of samples for each metric.

\begin{table}[h]
	\scriptsize
	\centering
	\begin{tabular}{l|lll}
		\toprule[1pt]
		\textbf{Task} & \makecell[c]{\textbf{Dialogue}\\\textbf{Summarization}}& \makecell[c]{\textbf{Question}\\\textbf{Generation}}
		& \makecell[c]{\textbf{Reading}\\\textbf{Comprehension}} \\
		\hline
		Dataset & SAMSum & Molweni & Molweni \\
		\#Train & 14,732 & 20,873 & 20,873 \\
		\#Val & 818& 2,346 & 2,346 \\
		\#Test & 819 & 2,560 & 2,560\\
		Output Length & 23.44$\pm$12.72 & 7.05$\pm$2.02 &4.01$\pm$2.93 \\
		%Outpus Length(std) & 12.72 & 2.02 & 2.93\\
		%\#Speaker & 1,932 & 8,770 & 8,770\\
		\bottomrule[1pt]
	\end{tabular}
	%\begin{tabular}{p{0.8cm}rrrrrr}
	%	\hline
	%	\textbf{Dataset} & \textbf{\#Train} &\textbf{ \#Val} & \textbf{\#Test}  & \textbf{Avg} & \textbf{Std} &\textbf{\#Speaker} \\
	%	\hline
	
	%	SAMSum & 14,732 & 818 & 819 & 23.44 & 12.72 & 1,932\\
	%	Molweni &20.873 & 2,346 & 2,560 & 7.05& 2.02 & 8,770\\

	%	\hline
	%\end{tabular}
	\caption{A summary of tasks. \#Train, \#Val and \#Test refer to the number of samples in the datasets. Output length are statistics(avg$\pm$std) for the word counts.}% \#Speaker corresponds to the number of samples for change-one-name tests in Sec~\ref{sec:quatification}.}
	\label{tab:taskdata}
\end{table}

\textbf{Dialogue Summarization} outputs fluent and concise summaries covering the salient information in dialogues. We experiment with the SAMSum dataset~\cite{gliwa2019samsum} consisting of around 16k open-domain dialogues among two or more interlocutors. Rouge-2 F1~\cite{lin2004rouge} and BertScore F1~\cite{zhang2019bertscore}\footnote{We adopted microsoft/deberta-xlarge-mnli recommended by {https://github.com/Tiiiger/bert\_score} for BertScore.} are task-specific evaluation metrics. We consider genders to be consistent when switching names following~\citet{khalifa2021bag}. 
%The input to the model is a concatenation of speaker-utterance pairs and the output is a narrative summary. 
%to evaluate the generated summaries by comparing with the reference. 

\textbf{Question Generation} is to generate a question given an input dialogue and its corresponding answer span. We use Molweni dataset~\cite{li2020molweni} made up of around 10k task-oriented dialogues sampled from the Ubuntu Chat Corpus.
Similar to the question generation work based on SQuAD1.1, we extract (dialogue, answer, question) tuples from the original Molweni dataset and ignore unanswerable questions. Bleu~\cite{papineni2002bleu} and Rouge-L F1 are used for evaluations. % An answer is appended at the end of the corresponding dialogue as the model's input.

\textbf{Reading Comprehension} generates an answer by inputting a dialogue with a question. We use the Molweni dataset~\cite{li2020molweni} and ignore unanswerable questions as well. Bleu and Rouge-L F1 are also used for evaluations.

 %, so that it is comparable to sensitivity metrics~\footnote{In the official implementations online, some scores are macro-averaged among samples while others are micro-averaged.}.

\subsection{Implementation Details}

We use BART-large
%~\footnote{\url{https://huggingface.co/facebook/bart-large}} 
as our basic pre-trained model. We truncate inputs to the first 1024 tokens and the learning rate is $3e-5$ with weight decay equaling 0.01. The model is fine-tuned with batch size equaling 32 for 10 epochs. We evaluate the performance on $D_{va}$ after each epoch with Rouge-2 F1 or Bleu. The checkpoint with the highest score on $D_{va}$ is saved for testing. During the inference, we decode with no\_repeat\_ngram\_size=3, length\_penalty=1.0 and num\_beams=4. We search $\alpha$ and $\beta$ in \{1, 10, 20\} empirically and report results with the best validation performance. Specifically, $\alpha$ equals $1$. $\beta$ equals $1$ for reading comprehension and $10$ for the others.
Our experiments are done on a single RTX 2080Ti with 11G GPU memory.
Considering the GPU memory footprint, we set $K=2$, which is the same for Aug and FreAug for fair comparisons. %\JQ{hyperparameter}

We test online approaches with their corresponding test sets. For offline approaches, we focus on two sources of $P$. One is \textbf{in-distribution names} representing speaker names from the corresponding $D_{tr}$. The other is \textbf{all-possible names} with more than 117 thousand names\footnote{\url{https://data.world/arunbabu/gender-by-names}}, which can reflect the models' performances in complicated real scenarios. 
For approaches with sampling operations, we construct data with 3 different random seeds. Results are averaged over the number of runs.
%, and 5 for testing, i.e., $T=5$ in Sec.~\ref{sec:quatification}

%The last one is \textbf{frequent names} which will be introduced in Section~\ref{sec:dda}.
%Other specific groups of speaker will be mentioned in Section~\ref{sec:unfairness}.

\section{Results}
\label{sec:results}

We show performances of approaches first, followed by ablation studies and  evaluations.
Then, we take a closer look at offline approaches, which show the inherent capability of models, with multi-faceted analysis. Hyper-parameter search and case studies are in Appendix~\ref{sec:app-hyper} and \ref{app:casestudy}.% Appendixes.%performing on different groups of people. 
%\footnote{Since we mainly focus on the differences of generation results which can be easily reflected by evaluation metrics, we didn't carry out human evaluations in this work.}
%At last, we show the  of our approach. %do ablation studies of our approach in the aspect of hyper-parameter search.

\subsection{Performance of Offline Approaches}
\label{sec:mda}
% original test set, in-distribution test set, all names test set

% It improves R2 with 0.17\% which is not consistent with the 0.40\% improvements in~\citet{liu2021controllable}. We think this is mainly due to the stochastic sampling operation of doing data augmentation.
%Our results are more reliable, since the results in our work are averaged over three times with training data augmented with different random seeds while \citet{liu2021controllable} didn't mention how many runs they took. 

The performance on the original test sets is shown in Table~\ref{tab:mdresults-vanilla}. Emb only outperforms Vanilla on question generation and Aug only makes little improvements over Vanilla on dialogue summarization. %and even drops on the other two tasks.
Our approach {Ins} makes consistent improvements, {performing best among offline approaches}.

\begin{table}[h]
	\scriptsize
	\centering
	\begin{tabular}{l|cccccccc}
		\toprule[1pt]
		& \multicolumn{2}{l}{\makecell[c]{\textbf{Dialogue}\\\textbf{Summarization}}}& \multicolumn{2}{c}{\makecell[c]{\textbf{Question}\\\textbf{Generation}}}
		& \multicolumn{2}{c}{\makecell[c]{\textbf{Reading}\\\textbf{Comprehension}}}  \\
		
		{\textbf{Approach}}  & \textbf{R2} & \textbf{BertS} & \textbf{Bleu} & \textbf{RL} & \textbf{Bleu} & \textbf{RL}\\
		\hline
		Vanilla & 28.12 &  75.09 & 18.57 & 56.04& 28.42&73.33  \\
		Emb & 28.12 & 75.14 & 19.97 & 56.83 & 26.35 & 69.31 \\
		Aug  & 28.29 &  75.26 & 18.53 & 55.56&27.09 & 71.88 \\

		%Cross & \\
		Ins$\star$  & \textbf{28.97} & \textbf{75.63} & \textbf{20.26} & \textbf{56.85} & \textbf{29.44} & \textbf{74.03}\\

		\bottomrule[1pt]
	\end{tabular}
	\caption{Performances(\%) of offline approaches on the original test set. Vanilla refers to the baseline that simply fine-tuned the basic pre-trained model on the original dataset for different tasks. $\star$ marks our approach.}
	\label{tab:mdresults-vanilla}
\end{table}

Results with sensitivity scores are in Table~\ref{tab:mdresults-main}. 
Emb fails to generate more insensitive results, especially for question generation.
{Aug} doesn't make promising improvements on outputs' quality over Vanilla, but it {reduces the sensitiveness of models} across different test sets and tasks.
Ins leads to better results on randomly augmented training data with different random seeds, significantly outperforming Aug. %with p-value<0.05.
In a word, Ins achieves the best performance among offline approaches. %, achieving the state-of-the-art results}.

By comparing the results in Table~\ref{tab:mdresults-main} horizontally, in-distribution names perform better than all-possible names on dialogue summarization, whereas results are opposite on the others. Speaker names in SAMSum are mostly real and popular names, while names in Molweni are online nicknames containing unknown words, such as ``zykotick9''. All-possible names contain a large proportion of real names, and a small proportion of names never seen during pre-training which can be regarded as nicknames. In this way, we can observe that the difficulty of modeling names for a model is ``SAMSum in-distribution  $<$ all-possible $<$ Molweni in-distribution''. In other words, models perform better on more popular names, which is in accord with the success of Fre in Sec.~\ref{sec:onlineapproach}.

\begin{table}[t]
	\scriptsize
	\centering
	\begin{subtable}{\linewidth}
		\scriptsize
		\centering
		\begin{tabular}{p{0.9cm}|p{0.36cm}p{0.36cm}p{0.36cm}p{0.36cm}|p{0.36cm}p{0.36cm}p{0.36cm}p{0.38cm}}
			\toprule[1pt]
			%	\textbf{Approach} & \textbf{R2} & \textbf{BertS} & \textbf{D-R2} & \textbf{R-R2} & \textbf{D-BertS} & \textbf{R-BertS}  \\
			& \multicolumn{4}{c|}{\textbf{R2}} & \multicolumn{4}{c}{\textbf{BertScore}} \\
			%\cline{2-9}
			\textbf{Approach}& - & S$\downarrow$  & R$\downarrow$ & D$\downarrow$& - & S$\downarrow$ & R$\downarrow$& D$\downarrow$  \\

			\hline
			\multicolumn{9}{l}{\textit{In-distribution Names}}\\
			Vanilla & 27.66 & 31.24  & 13.98 & 5.51 &74.90&11.80&6.41& 2.49\\
			Emb & 27.63 & 29.39& 13.21 & 5.20 &74.91 &11.29 & 6.26& 2.43 \\
			Aug & 27.82 &27.35  & 12.33 & 4.86&74.95& 10.42 & 5.77 &2.57 \\

			%Cross & \\
			Ins$\star$  &  \underline{\textbf{28.79}} & \underline{\textbf{21.36}} & \underline{\textbf{9.50}} & \underline{\textbf{3.82}}& \textbf{75.48}&\underline{\textbf{7.94}} & \underline{\textbf{4.32}}& \underline{\textbf{1.71}} \\
			
			\hline
			\multicolumn{9}{l}{\textit{All-possible Names}}\\
			Vanilla & 27.19 & 33.10& 14.64 & 5.72 &74.83&12.26 & 6.66 & 2.60\\
			Emb & 27.22 & 31.38 & 13.59 & 5.30&74.89 &12.03 & 6.63 & 2.55\\
			Aug &27.50 &28.17 & 12.56 & 4.97 &74.96 &10.56 & 5.76 & 2.25 \\

			%Cross & \\
			Ins$\star$  &  \underline{\textbf{28.44}} & \underline{\textbf{25.37}}  & \textbf{11.58} & \textbf{4.62}&\textbf{75.38} &\underline{\textbf{9.38}} & \textbf{5.22}& \textbf{2.05} \\
			
			\bottomrule[1pt]
		\end{tabular}
		\caption{Dialogue Summarization}
		\label{tab:mdresults-ds}
	\end{subtable}
	
	\begin{subtable}{\linewidth}
		\scriptsize
		\centering
		\begin{tabular}{p{0.9cm}|p{0.36cm}p{0.36cm}p{0.36cm}p{0.36cm}|p{0.36cm}p{0.36cm}p{0.36cm}p{0.38cm}}
			\toprule[1pt]
			
			%\textbf{Approach} & \textbf{Bleu} & \textbf{RL} & \textbf{D-Bleu} & \textbf{R-Bleu} & \textbf{D-RL} & \textbf{R-RL}  \\
			& \multicolumn{4}{c|}{\textbf{Bleu}} & \multicolumn{4}{c}{\textbf{RL}} \\
			%\cline{2-9}
			\textbf{Approach}& - & S$\downarrow$ & R$\downarrow$ & D$\downarrow$ & - & S$\downarrow$ & R$\downarrow$ & D$\downarrow$ \\
			
			\hline
			\multicolumn{9}{l}{\textit{In-distribution Names}}\\
			Vanilla & 18.48 &  34.80& 11.96 & 5.06&{57.14}&14.94&14.19 & 5.74\\
			Emb & 19.00 &38.24& 13.76 & 5.79 & 57.31 &17.55& 16.85 & 6.82 \\
			Aug & 17.89 & 26.24  & 8.22& {3.52}&56.26& 12.04& {11.35} & {4.69} \\

			%Cross & \\
			Ins$\star$ &  \textbf{19.58 }&\underline{\textbf{16.90}} & \underline{\textbf{5.53}} & \underline{\textbf{2.35}}& \textbf{57.47} &\underline{\textbf{7.83}} & \underline{\textbf{8.09}} &\underline{\textbf{3.35}}\\

			\hline
			\multicolumn{9}{l}{\textit{All-possible Names}}\\
			Vanilla & 18.56 & 29.64 & 10.04 & 4.26& {57.38}&12.98& 11.88 & 4.90 \\
			Emb & 18.70 & 35.52& 12.55 & 5.27 &57.28 &16.05 & 15.26 & 6.20 \\
			Aug & 17.81 &  23.09 & {7.15} & {3.06}&56.08&10.66& {9.64} & {4.03} \\

			%Cross & \\
			Ins$\star$ &  \textbf{19.57} & \underline{\textbf{14.65}} & \underline{\textbf{4.41}} & \underline{\textbf{1.90}} &\textbf{57.49}&\underline{\textbf{6.96}} & \underline{\textbf{6.58}} & \underline{\textbf{2.78}}\\
			
			\bottomrule[1pt]
		\end{tabular}
		\caption{Question Generation}
		\label{tab:mdresults-qg}
	\end{subtable}
	
	\begin{subtable}{\linewidth}
		\scriptsize
		\centering
		\begin{tabular}{p{0.9cm}|p{0.36cm}p{0.36cm}p{0.36cm}p{0.36cm}|p{0.36cm}p{0.36cm}p{0.36cm}p{0.38cm}}
			\toprule[1pt]
			%\textbf{Approach} & \textbf{Bleu} & \textbf{RL} & \textbf{D-Bleu} & \textbf{R-Bleu} & \textbf{D-RL} & \textbf{R-RL} \\
			& \multicolumn{4}{c|}{\textbf{Bleu}} & \multicolumn{4}{c}{\textbf{RL}} \\
			%\cline{2-9}
			\textbf{Approach}& - & S$\downarrow$ & R$\downarrow$& D$\downarrow$  & - & S$\downarrow$  & R$\downarrow$& D$\downarrow$ \\
			\hline
			\multicolumn{9}{l}{\textit{In-distribution Names}}\\
			Vanilla &28.34 & 54.98 & 6.54 & 2.83 &73.07&7.54& 9.69 & 4.17 \\
			Emb & 25.80 &57.78& 7.17 & 3.13 & 69.29 &9.83& 12.30 & 5.31 \\
			Aug & 27.07 & 55.96& 6.04 & 2.62 & 72.11&8.14 & 10.42&  4.50 \\

			Ins$\star$ & \underline{\textbf{29.31}} &\underline{\textbf{52.03}}& \underline{\textbf{4.53}} & \underline{\textbf{1.97}} & \textbf{74.04} &\underline{\textbf{5.65}} & \underline{\textbf{7.66}} & \underline{\textbf{3.32}}\\
			
			\hline
			\multicolumn{9}{l}{\textit{All-possible Names}}\\
			Vanilla &  28.56 & 53.94& 5.39 & 2.34& 73.60&6.39 & 8.21  & 3.53\\
			Emb &25.99 &  56.22 & 5.11 & 2.21&69.59&7.29 & 8.60 & 3.69\\
			Aug & 27.12 & 54.72 & 5.15 & 2.23& 72.23& 6.39 & 8.29  & 3.58\\
			
			Ins$\star$ &  \underline{\textbf{29.34}} &\underline{\textbf{51.38}}  & \underline{\textbf{3.66}} & \underline{\textbf{1.59}}& \textbf{74.35}&\underline{\textbf{4.62}} & \underline{\textbf{6.15}} & \underline{\textbf{2.64}}\\
			
			\bottomrule[1pt]
		\end{tabular}
		\caption{Reading Comprehension}
		\label{tab:mdresults-rc}
	\end{subtable}
	\caption{Performances(\%) of offline approaches. ``-'' is the original metric. S, D and R are shorted for the sensitivity metrics. Scores significantly better than all the baselines with p-value$<$0.05 are underlined.}	
	\label{tab:mdresults-main}
\end{table}

\subsection{Performance of Online Approaches}
\label{sec:onlineapproach}
% replaced test set

The results of online approaches are in Table~\ref{tab:ddresults-main}.

\begin{table}[h]
	\scriptsize
	\centering
	\begin{subtable}{\linewidth}
		\scriptsize
		\centering
		\begin{tabular}{p{0.9cm}|p{0.36cm}p{0.36cm}p{0.36cm}p{0.36cm}|p{0.36cm}p{0.36cm}p{0.36cm}p{0.38cm}}
			\toprule[1pt]
			
			%\textbf{Approach} & \textbf{Bleu} & \textbf{RL} & \textbf{D-Bleu} & \textbf{R-Bleu} & \textbf{D-RL} & \textbf{R-RL}  \\
			
			 & \multicolumn{4}{c|}{\textbf{R2}} & \multicolumn{4}{c}{\textbf{BertScore}} \\
			%\cline{2-9}
			\textbf{Approach}& - & S$\downarrow$  & R$\downarrow$ & D$\downarrow$& - & S$\downarrow$  & R$\downarrow$& D$\downarrow$ \\
			\hline
			%Vanilla & 28.12 & 75.09 & - &-&-&-\\
			ID & 26.97 & - & - &-&74.26&-&-&-\\
			Fre & {28.55} & 25.17 & 11.31 & 4.50 &74.24&9.77 & 5.30 & 2.09 \\
			FreAug & 27.86 & 25.03 & {11.09} & {4.39}&75.02 &9.58 & {5.12}& 2.02 \\
			
			FreIns$\star$ & \underline{\textbf{28.73}} &\underline{\textbf{17.25}} & \underline{\textbf{7.66}} & \underline{\textbf{3.14}}& \underline{\textbf{75.53}}&\underline{\textbf{6.39}}& \underline{\textbf{3.43}} & \underline{\textbf{1.38}} \\
			\bottomrule[1pt]
		\end{tabular}
		\caption{Dialogue Summarization}
		\label{tab:ddresults-ds}
	\end{subtable}
	
	\begin{subtable}{\linewidth}
		\scriptsize
		\centering
		\begin{tabular}{p{0.9cm}|p{0.36cm}p{0.36cm}p{0.36cm}p{0.36cm}|p{0.36cm}p{0.36cm}p{0.36cm}p{0.36cm}}
			\toprule[1pt]
			%\textbf{Approach} & \textbf{Bleu} & \textbf{RL} & \textbf{D-Bleu} & \textbf{R-Bleu}& \textbf{D-RL} & \textbf{R-RL}  \\
			 & \multicolumn{4}{c|}{\textbf{Bleu}} & \multicolumn{4}{c}{\textbf{RL}} \\
			%\cline{2-9}
			\textbf{Approach}& - & S$\downarrow$ & R$\downarrow$ & D$\downarrow$ & - & S$\downarrow$  & R$\downarrow$& D$\downarrow$ \\
			\hline
			%Vanilla & 18.57 & 56.40 & - &-&-&-\\
			ID & {19.21} & - & - &-&56.49&-&-&-\\
			Fre & {18.96} & 18.44  & 5.51 & 2.35&{57.10}&8.35& 7.23  & 3.04 \\
			FreAug &  18.52 & 16.01& 4.92 & 2.14& 57.06 & 7.05& 6.50& 2.76 \\
			FreIns$\star$ & \textbf{19.71} &\underline{\textbf{10.09}} & \underline{\textbf{3.12}} & \underline{\textbf{1.35}}& \textbf{57.29} &\underline{\textbf{4.48}} & \underline{\textbf{4.19}} & \underline{\textbf{1.80}}\\
			\bottomrule[1pt]
		\end{tabular}
		\caption{Question Generation}
		\label{tab:ddresults-qg}
	\end{subtable}
	
	\begin{subtable}{\linewidth}
		\scriptsize
		\centering
		\begin{tabular}{p{0.9cm}|p{0.36cm}p{0.36cm}p{0.36cm}p{0.36cm}|p{0.36cm}p{0.36cm}p{0.36cm}p{0.36cm}}
			\toprule[1pt]
			%\textbf{Approach} & \textbf{Bleu} & \textbf{RL} & \textbf{D-Bleu} & \textbf{R-Bleu} & \textbf{D-RL} & \textbf{R-RL} \\
			& \multicolumn{4}{c|}{\textbf{Bleu}} & \multicolumn{4}{c}{\textbf{RL}} \\
			%\cline{2-9}
			\textbf{Approach}& - & S$\downarrow$  & R$\downarrow$ & D$\downarrow$& - & S$\downarrow$  & R$\downarrow$ & D$\downarrow$\\
			\hline
			%Vanilla & 28.42 & 73.33 & - &-&-&-\\
			ID & 28.46 & - & - &-&73.62&-&-&-\\
			Fre & 27.35 & 54.55& 3.77 & 1.63& 73.56 &4.95& 6.05 & 2.61  \\
			FreAug & 27.92 & 52.67 & 3.28& 1.42 &73.67&4.24& 5.63& 2.43  \\
			FreIns$\star$ &  \textbf{29.03} &\textbf{52.28}  & \textbf{2.66}& \textbf{1.15}& \textbf{74.59}&\underline{\textbf{3.28}}  & \underline{\textbf{4.51}}& \underline{\textbf{1.95}}\\
			\bottomrule[1pt]
		\end{tabular}
		\caption{Reading Comprehension}
		\label{tab:ddresults-rc}
	\end{subtable}
	\caption{Performances(\%) of online approaches.}%Scores significantly better than the baselines with p-value$<$0.05 are underlined. }	
	\label{tab:ddresults-main}
\end{table}

All speaker names will be normalized into fixed code names in {ID}, so that the test set for ID is changeless for each sample and the sensitivity scores are actually 0.0. Unfortunately, its quality scores lag behind Ins and even drop dramatically on dialogue summarization. Thus, it's not recommended to be a necessary data pre-processing step.  % The differences among code names is only a number, making them more indistinguishable by models. Moreover,  % However, comparing the quality metrics with other approaches, the  % Some latent but important characteristics in names are neglected such as gender.

{Fre} makes some improvements on R2 for dialogue summarization by comparing with the vanilla model, which is consistent with the results in~\cite{khalifa2021bag}, whereas the drops in BertScore were not mentioned in their work. The sensitivity scores are lower than those for offline approaches in Table~\ref{tab:mdresults-main}. To better understand the gains of Fre, we further test the vanilla model with the same test sets replaced by frequent names. It achieves similar performance on Rouge-2 (28.18) and BertScore (75.13) with the vanilla model. The sensitivity score D-BertS is 2.24, which is lower than 2.49 of Vanilla in Table~\ref{tab:mdresults-main}. It shows that {the advantages of Fre not only come from using the group of frequent names} that are easier for a model to understand, {but also from doing fine-tuning with this group of names}. FreAug doesn't improve the outputs' quality consistently, but reduces the sensitivity scores. %This is the same as the comparisons between Aug and Vanilla. 

{FreIns} performs the most insensitively with better generation quality among online approaches.

\subsection{Ablation Study}

Ablation studies of our full approach Ins are in Table~\ref{tab:ablation}.
Aug is regarded as an ablation representing the model trained without any auxiliary losses.
Both insensitivity losses outperform Aug with using $\mathcal{L}_{dh}$ topping the rank on most metrics, showing that penalizing differences on the decoder hidden states has more direct effects on the outputs.
Combining both losses induces more performance gains.

% DH and CA refer to the model trained only with $\mathcal{L}_{ca}$ or $\mathcal{L}_{dh}$ respectively, and 

\begin{table}[th]
	\scriptsize
	\centering
	\begin{tabular}{p{0.99cm}|p{0.3cm}cp{0.3cm}cp{0.3cm}c}
		\toprule[1pt]
		& \multicolumn{2}{l}{\makecell[c]{\textbf{Dialogue}\\\textbf{Summarization}}}& \multicolumn{2}{c}{\makecell[c]{\textbf{Question}\\\textbf{Generation}}}
		& \multicolumn{2}{c}{\makecell[c]{\textbf{Reading}\\\textbf{Comprehension}}}  \\
		{\textbf{Approach}}  & \textbf{BertS} & \textbf{D-BertS$\downarrow$} & \textbf{Bleu} & \textbf{D-Bleu$\downarrow$} & \textbf{Bleu} & \textbf{D-Bleu$\downarrow$}\\
		\hline
		Ins & \textbf{75.48} & \textbf{1.71} & 19.48 & \textbf{2.35} & \textbf{29.31} & \textbf{1.97} \\
		-w/o $\mathcal{L}_{ca}$ & 75.43 & 1.85 & \textbf{19.71} & 2.47 & 29.03 & 2.19 \\
		-w/o $\mathcal{L}_{dh}$& 74.89 & 2.27 & 18.40 & 3.01 & 28.42 & 2.04 \\
		Aug& 74.95 & 2.57 & 17.89 & 3.52 & 27.07 & 2.62 \\
		
		\bottomrule[1pt]
	\end{tabular}
	\caption{Ablations(\%) of the full approach Ins.}%with the in-distribution test setting.}
	\label{tab:ablation}
\end{table}

\subsection{Human Evaluation}

\begin{figure}[t]
	\centering
	\includegraphics[width=0.95\columnwidth]{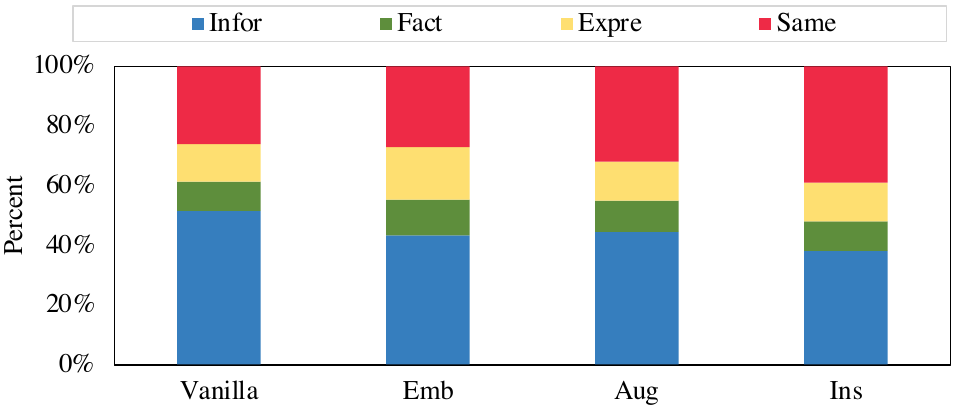}
	\caption{Human evaluation for difference types.} %in the summaries.}
	%\underline{Divergent contents} are underlined in the generated summaries.}
	\label{fig:humaneval}
\end{figure}

Taking dialogue summarization as an example, we did human evaluation to further prove the improvement on sensitivity by sampling 200 pairs of generations for each offline approach and asked three proficient English speakers from Asia to label each case out of 4 choices by selecting the primary one that makes the generations distinct: %  distinguishing one out of 4 choices:
%among 4 choices by choosing the major one that makes generations distinct: 
\textbf{Infor}mation difference means both outputs contain different information or keywords. 
\textbf{Fact}ual difference refers to different matchings between speakers and events.
%\textbf{Spea}ker reorder represents different orders of juxtaposed names.
\textbf{Expre}ssion difference is that the outputs have minor differences, such as capitalization and different orders of juxtaposed names.
\textbf{Same} represents the identical outputs.
The results are in Fig.~\ref{fig:humaneval} with 0.64 Kappa score, indicating substantial agreement. We can see that content distinction is the primary difference type. Ins generates less distinct contents and more identical results, outperforming the baselines.

\subsection{Sensitivity among Name Groups}
\label{sec:unfairness}
\begin{figure}[t]
	
	\begin{minipage}[t]{\linewidth}
		\centering
		\subfloat[Sensitivity among different popularity groups.]{
			\includegraphics[scale=0.48]{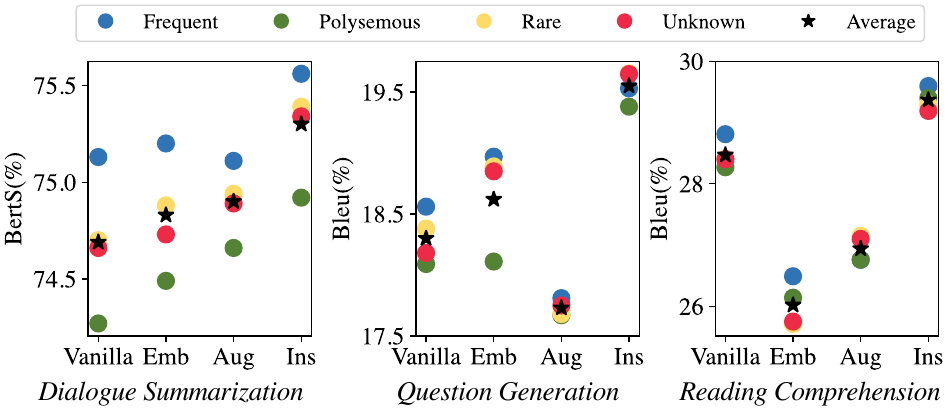}
			\label{fig:groups-frequency}
			%\caption{fig1}
		}%
	\end{minipage}%
	
	\begin{minipage}[t]{\linewidth}
		\centering
		\subfloat[Sensitivity among different racial groups.]{
			\includegraphics[scale=0.48]{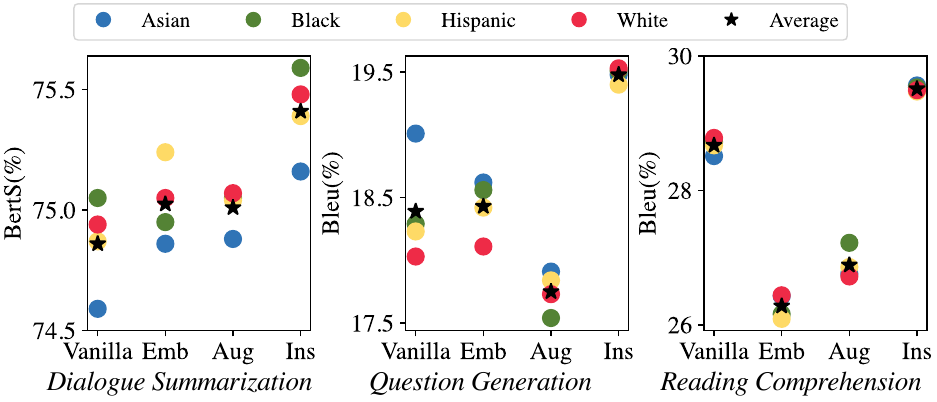}
			\label{fig:groups-racial}
		}%
	\end{minipage}
	\centering
	\caption{Sensitivity among names within different groups. The scores are the higher the better and more centralized dots for each approach represent better insensitivity among groups. }% The right figures shows intra-group fairness and the scores are expected to be lower.}	
	\label{fig:groups}
\end{figure}

% divide by name occurance
We collect specific groups of names in terms of popularity and race and show differences in the quality performances on test sets constructed with corresponding names. 
%People from the same race are likely to have more communication. A severe ethical issue will show up if there exists unfairness among the different races. Although the popularity of names for speakers is stochastic for each dialogue, the model still has the potential facing dramatic performance drops caused by ``unpopular'' names from speakers who lead the dialogue. This may also accelerate the extinction of rare names, together with their cultures. 
%We illustrate the performances of different name groups under each approach 
%The results are illustrated in Fig.~\ref{fig:groups} and Figure~\ref{fig:groups2}. 
The sensitivity among different groups for each method are reflected by the scattering of dots vertically in Fig.~\ref{fig:groups}.% and Fig.~\ref{fig:groups2}. %Task-specific metrics on corresponding test sets show the fairness among groups, i.e. inter-group fairness. %The fairness within groups, i.e. intra-group fairness, is compared by sensitivity metrics.

%\begin{figure}[t]
%	\centering
%	\begin{minipage}[t]{\linewidth}
%		\centering
%		\subfloat[Dialogue Summarization]{
%			\includegraphics[scale=0.48]{samsum.pdf}
%			%\caption{fig1}
%		}%
%	\end{minipage}%
%	
%	\begin{minipage}[t]{\linewidth}
%		\centering
%		\subfloat[Question Generation]{
%			\includegraphics[scale=0.48]{molweni.pdf}
%			\label{fig:groups-mowelni}
%		}%
%	\end{minipage}
%	\centering
%	\caption{Unfairness among different groups of names.}	
%	\label{fig:groups}
%\end{figure}

%\label{sec:unfairnessstatistical}

\textbf{Name groups by popularity and usage:} We define 4 groups. \textbf{Frequent} including words frequently and solely used as human names is mentioned before. \textbf{Polysemous} represents words frequently used but not specialized for human names, such as June and Florida. \textbf{Rare} is names with low occurrence times like Paderau. \textbf{Unknown} names are similar to random strings from a model's perspective since they haven't been exposed to the model. The last three groups are collected by counting occurrences
of all-possible names in the pre-training corpus of BART. We select 200 names for each group (More details are in Appendix~\ref{sec:app-names}).
%100 males and 100 females for each group. 
%\JQ{polysemous}

According to Fig.~\ref{fig:groups-frequency}, we can see that models usually perform poorly on Polysemous, % for inter-group fairness, %and intra-group fairness, 
even worse than Rare and Unknown.  The daily meanings dominate the representation of this word and confuse the model. 
Frequent generally outperforms other groups.
%on both generation quality and on sensitivity scores. For example, R-BertS of the vanilla model tested with Frequent is only 5.70\%, which is not only significantly lower than other groups (Polysemous 7.79\%, Rare 6.69\%, Unknown 6.79\%), but also better than Vanilla with 6.41\% or 6.66\% in Table~\ref{tab:mdresults-main}. 
We conclude that 
words frequently and uniquely used as names that result in more specialized embeddings in pre-trained models and perform better. 
Moreover, comparing the sensitivity among different approaches, Ins outperforms the baselines in most cases except Aug. It achieves more centralized dots due to the performance reduction on the dominant groups or even all groups, showing that models tend to overfit with augmented data without our losses.
%When armed with our losses, the model understands the similarities and differences among augmented samples better and thus performs better with all groups
%Taking dialogue summarization in Fig.~\ref{fig:groups-frequency} as an example, Aug achieves more centralized dots mainly due to the performance decreases on the dominant groups, indicating that the model tends to overfit on these names. With the guidance of our losses, the model take better advantage of the augmented data  
To recap, Ins results in consistent improvements over Vanilla among different tasks compared with other baselines.
%Aug and Ins result in more fair performances. %with Ins topping the rank. 
% divide by racial

\textbf{Name groups by races:} Names from different races are from~\citet{tzioumis2018demographic} by assigning each name to a race with the highest probability. 4 major groups\footnote{Other groups are empty after this assigning operation with \citet{tzioumis2018demographic}'s name list.} are gathered, including Non-Hispanic \textbf{White}, \textbf{Hispanic} or Latino, Non-Hispanic \textbf{Black} or African American, and Non-Hispanic \textbf{Asian} or Native Hawaiian or Other Pacific Islander. % containing 2,873, 424, 64 and 561 names respectively. 
To avoid the influence of the various number of names, we select the most frequent 50 names in each group and show the results in Fig.~\ref{fig:groups-racial}.
All of the approaches show discrimination against Asian in dialogue summarization. %according to inter-group results. %, and superiorities on White with lowest intra-group sensitivity. 
Emb, Aug and Ins improve the insensitivity among different races compared with Vanilla, and Ins is better with the guarantee on quality. We consider to introduce special designs on demographic features in the future.

%\begin{figure}[t]
	%\centering
	%\begin{minipage}[t]{\linewidth}
	%	\centering
	%	\subfloat[Dialogue Summarization]{
	%		\includegraphics[scale=0.5]{samsum-racial.pdf}
	%		%\caption{fig1}
	%	}%
	%\end{minipage}%
	%
	%\begin{minipage}[t]{\linewidth}
	%	\centering
	%	\subfloat[Question Generation]{
	%		\includegraphics[scale=0.5]{molweni-racial.pdf}
	%		\label{fig:groups-mowelni}
	%	}%
	%\end{minipage}
%	\centering
%	\includegraphics[scale=0.5]{racial.pdf}
%	\caption{Sensitivity among different racial groups.  }	
%	\label{fig:groups2}
%\end{figure}
\subsection{Sensitivity on an Individual Speaker}

We can also only change the name of a single speaker each time to analyze fine-grained sensitivity.
The results of offline approaches for dialogue summarization are shown in Table~\ref{tab: change-one-name} (see more in Appendix~\ref{sec:app-results}).
The sensitivity scores are lower than the ones in Table~\ref{tab:mdresults-main}. It seems that the sensitivity of models is proportional to the amount of changes in test samples, i.e., whether changing all speaker names (change-all-name) or only one speaker name (change-one-name). However,
%by analyzing the relations of sample-wise and speaker-wise sensitivity scores for each sample, 
it's not always true and changing one name can be more sensitive than changing all names. Taking the results from Ins as an example, around 52.01\% samples have speakers whose change-one-name D-BertS is higher than the corresponding changel-all-name one. Over 34.80\% of the change-one-name D-BertS averaged by speakers from the same dialogue is also higher than the change-all-name D-BertS.

\begin{table}[h]
	\scriptsize
	\centering
	\begin{tabular}{p{0.9cm}|p{0.36cm}p{0.36cm}p{0.36cm}p{0.36cm}|p{0.36cm}p{0.36cm}p{0.36cm}p{0.38cm}}
		\toprule[1pt]
		
		%\textbf{Approach} & \textbf{Bleu} & \textbf{RL} & \textbf{D-Bleu} & \textbf{R-Bleu} & \textbf{D-RL} & \textbf{R-RL}  \\
		
		 & \multicolumn{4}{c|}{\textbf{R2}} & \multicolumn{4}{c}{\textbf{BertScore}} \\
		%\cline{2-9}
		\textbf{Approach}& - & S$\downarrow$  & R$\downarrow$ & D$\downarrow$& - & S$\downarrow$  & R$\downarrow$& D$\downarrow$ \\
		\hline
		\multicolumn{7}{l}{\textit{In-distribution Names}}\\
		Vanilla & 27.29 &25.53 & 11.05& 4.42& 74.64 &9.65&5.19 & 2.05\\
		Emb & 27.41 &24.20 & 10.87 & 4.33& 74.90 &9.49 & 5.29& 2.09 \\
		Aug &
		{27.51} & 22.24 & {9.89} & {3.96} &74.83&8.50 & 4.67 & {1.85}\\
		Ins$\star$ & \textbf{28.70} &\underline{\textbf{16.54}}  & \textbf{7.19} & \textbf{2.92}& \textbf{75.44}&\underline{\textbf{6.11}}& \textbf{3.18}& \textbf{1.28} \\
		
		\hline
		\multicolumn{7}{l}{\textit{All-possible Names}}\\
		Vanilla &
		27.32& 23.77  & 11.07 & 4.45&74.81 &9.61 &5.15& 2.04  \\
		Emb & 27.26 & 24.98 & 10.68 & 4.25& 75.30&9.57 & 5.16& 2.02\\
		Aug  &
		27.36 & 22.73 & {10.04} & {4.03} &74.86&8.56 & 4.69 & 1.87\\
		Ins$\star$  & \textbf{28.38} &\underline{\textbf{18.65}} & \textbf{8.12}& \textbf{3.29}  &\textbf{75.35} &\underline{\textbf{6.89}}& \textbf{3.75} & \textbf{1.50} \\
		
		\bottomrule[1pt]
	\end{tabular}
	\caption{Dialogue summarization results(\%) of offline approaches for sensitivity on an individual speaker.}
	\label{tab: change-one-name}
\end{table}

\begin{figure}[h]
	\centering
	\includegraphics[scale=0.55]{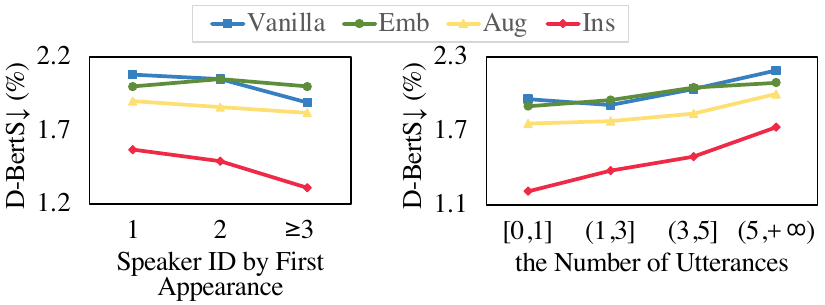}
	
	\caption{Change-one-name sensitivities on different speaker features for dialogue summarization.}%with all-possible names
	\label{fig:trends}
\end{figure}

We further show the trends between speaker features and their sensitivity scores in Fig.~\ref{fig:trends}. Names are more sensitive and thus crucial for speakers at the start of a dialogue or with more utterances, deserving attention for further improvements.

\section{Related Work}

\textbf{Entity/Name Bias in Narrative Texts}: 
Previous work on entity biases shows that pre-trained language models are sensitive to changes in narrative text. Some works~\cite{zhang2018graph,zhang-etal-2017-position,wang-etal-2022-rely} for relation extraction mask entities in the context to prohibit learning spurious features between entities and relations. \citet{yan2022robustness} analyzes the robustness of models by entity renaming on reading comprehension. They all consider different kinds of entities, such as person and organization. However, the entities have the potential to be grounded in real life~\cite{smith2021hi}, and the background knowledge of these entities may be necessary for understanding. Besides, the context and the entities cannot always be well-separated, especially persons~\citet{yan2022robustness}. Thus, masking and switching operations are not always suitable for these entities.
In our work, we focus on speakers that are not grounded.  %don't consider entities mentioned in dialogues and 

Names that are not grounded have also been studied. Information such as age, gender and race can be reflected by a given name to some extent~\cite{girma2020black}, while models learned with statistical features may make wrong predictions about specific persons or bring unexpected stereotypes~\cite{bertrand2004emily}. \citet{romanov2019s} takes occupation classification as an example and discourages the model to predict an individual's occupation depending on his/her name. \citet{wang2022measuring} presents that machine translation models perform poorly on female names when translating into languages with grammatical gender and also have sentiment bias caused by names with sentiment-ambiguous words. Samples in all these works only have a single name each, while multiple speaker names are entangled in a single dialogue.

%entities:

%[2020EMNLP]"You are grounded!": Latent Name Artifacts in Pre-trained Language Models

%[2022NAACL]On the Robustness of Reading Comprehension Models to Entity Renaming

%[2022NAACL] Should We Rely on Entity Mentions for Relation Extraction? Debiasing Relation Extraction with Counterfactual Analysis

%entities are likely to be grounded. speakers  are more ideal research target to 
%classification tasks , sequence generation tasks.

%names:

%[2022ACL]Measuring and Mitigating Name Biases in Neural Machine Translation, gender biases, sentiment biases

%[2019NAACL] What's in a Name? Reducing Bias in Bios without Access to Protected Attributes, occupation bias, protected attributes

%dialogues: more names in a single input

\textbf{Fairness of Dialogue Models}: 
Safety and fairness issues on generations from dialogue models are crucial for implementation in practice. Harmful differences in responses caused by different demographic personas are observed in well-known dialogue systems~\cite{sheng2021revealing,dinan2020queens}, including offensiveness, gender bias, race discrimination, etc. These unfairness phenomena also exist in dialogue systems without considering persons~\cite{liu2020does}, reflected by the politeness, sentiment, diversity and other aspects of a response. Recent work from~\cite{smith2021hi} shows dialogue models treat their conversation partner differently for different speaker names. Instead of analyzing differences in open-ended dialogue systems, we target on text generation tasks given dialogues and show that sensitivity/unfairness also exists among speakers.

%[2021]Revealing Persona Biases in Dialogue Systems.pdf:
%[2020COLING]Does Gender Matter? Towards Fairness in Dialogue Systems
%[2020]Queens are Powerful too: Mitigating Gender Bias in Dialogue Generation

%[2021]Hi, my name is Martha: Using names to measure and mitigate bias in generative dialogue models

%dialogue understanding tasks, multiple speakers with different genders or races are entangled.

\section{Conclusion}

This paper focuses on the speaker name sensitivity in the text generation from dialogues. We provide a classification for previous approaches, and propose the insensitivity losses to reduce the sensitivity while achieving favorable generation quality. Fair comparisons and comprehensive analysis are done among different approaches for evaluating the sensitivity quantitatively. 
More approaches targeting dialogue sensitivity issues are expected.

\section*{Limitations}

Our work has the following limitations:

First, we cannot generalize our conclusions to other languages that are dramatically different from English or more complicated multi-lingual scenarios without further experiments.

Second, we didn't consider any special designs on demographic features of names in our proposed approach. As shown in Sec.~\ref{sec:unfairness}, discrimination does exist among different groups. Although Ins outperforms other baselines overall, there is still room to improve insensitivity among different groups for tasks with longer outputs containing multiple speaker names. We hypothesize that demographic features of names can be added through a more dedicated data augmentation strategy.

\begin{figure}[h!]
	\centering
	\includegraphics[width=0.95\linewidth]{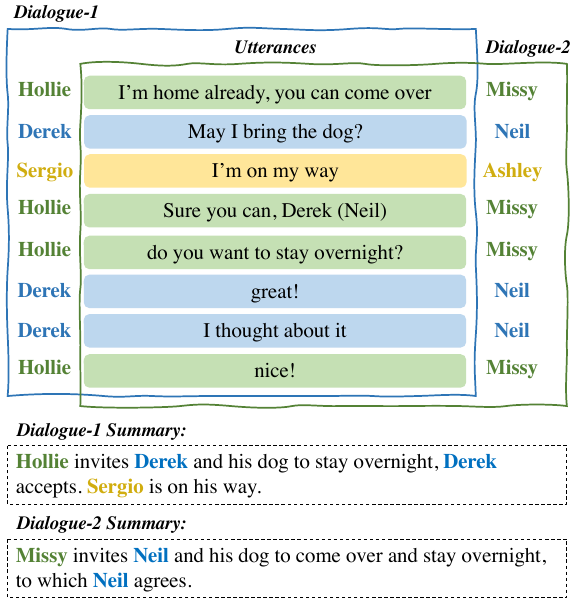}
	\caption{An example of dialogue summarization with outputs from ChatGPT.} 
	\label{fig:case-chatgpt}
\end{figure}

Third, our experimentation was restricted to the BART model in this paper. The reason is that among all the models that can be fine-tuned with our limited resources, including T5 and GPT-2, BART is still the best and the most popular, therefore we pick BART as the target of this study. Our intention is to devote the limited paper space to a more in-depth analysis of the problem using a range of tasks. Besides, it should be noticed that the speaker name sensitivity is still an issue with recent large pre-trained models, as shown in the example of dialogue summarization with outputs from ChatGPT in Fig.~\ref{fig:case-chatgpt}. The two summaries are expected to be the same, modulo speaker names. However, the third speaker (Sergio/Ashley) is not even mentioned in Summary-2.

We will try to address these limitations in the future.

\section*{Ethics Statement}

All of the name lists we adopted in this paper are borrowed from public websites (\url{https://www.ssa.gov}) and previous publications~\cite{tzioumis2018demographic, khalifa2021bag}. We considered only binary genders and four different racial groups, which are clearly incomplete for depicting all humans. Our work is mainly at drawing researchers' attention to the unfairness caused by speaker names in text generation tasks given dialogues. These demographic features are selected to shed light on this potential issue and our method is not restricted to any specific demographic groups.

\section*{Acknowledgments}
This work was generously supported by the CMB Credit Card Center \& SJTU
joint research grant, and Meituan-SJTU joint research grant.

\bibliography{acl23}
\bibliographystyle{acl_natbib}
%\newpage

%\newpage
%\clearpage
\appendix
%\section{A pilot experiment with human evaluation}
%\label{sec:appendix}

%We did a pilot experiment on the SAMSum dataset by with vanilla fine-tuned BART. By switching names, we collect two generations for each sample and names are changed back for comparisons. Among the 71.92\% samples with different pairs of outputs, we randomly sampled 200 pairs and asked two human annotators to label each case among 4 choices mentioned in Sec.5.5. 

%According to the results in Table~\ref{tab:pilot}, the inter-annotator agreement is almost perfect and the most severe unfairness is from content distinction. This inspired us to design the following loss function.

%\begin{table}[h]
%	\scriptsize
%	\centering
%	\begin{tabular}{crrrrr}
	%	\hline
%		\textbf{Annotator} & \textbf{Cont} &\textbf{Spea} & \textbf{Reas}  & \textbf{Expre} & \textbf{Kappa} \\
%		\hline
%		1 & 64.50 & 3.00 & 11.00 & 21.50 & \multirow{2}{*}{95.94}\\
%		2 & 64.50 & 3.00 & 13.00 & 19.50 & \\
		
%		\hline
%	\end{tabular}
%	\caption{The percentage of different labels(\%) and Kappa score(\%) between two annotators.}
%	\label{tab:pilot}
%\end{table}

\section{Illustration for Insensitivity Losses}
\label{sec:app-model}

\begin{figure*}[]
	\centering
	\includegraphics[scale=0.6]{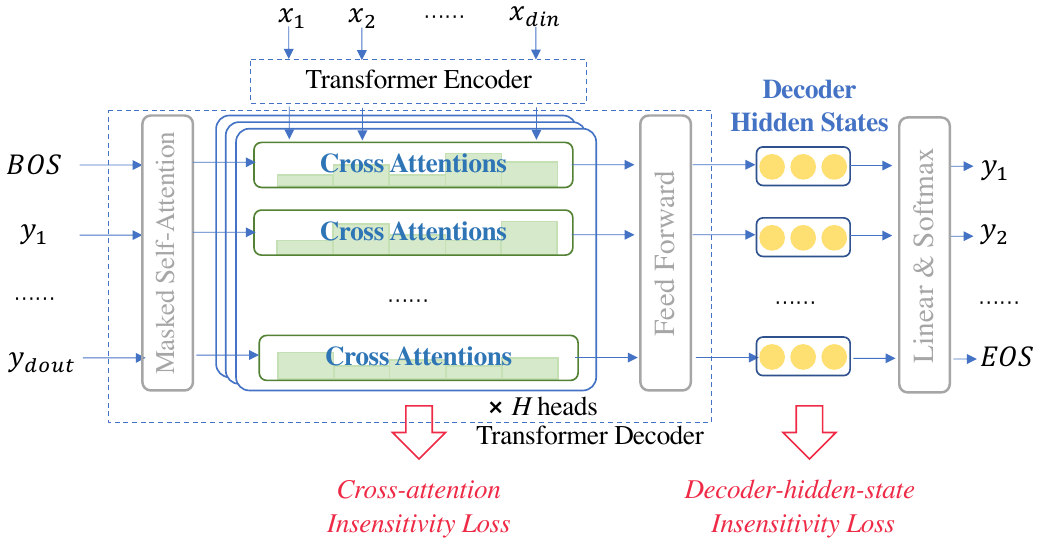}
	\caption{An illustration of insensitive losses. BOS and EOS are special tokens standing for the start and the end of the output.} %in the summaries.}
	%\underline{Divergent contents} are underlined in the generated summaries.}
	\label{fig:model}
\end{figure*}

Fig.~\ref{fig:model} depicts the positions of the cross attentions and the final decoder hidden states in the encoder-decoder Transformer model for a better understanding of our two insensitivity losses.

\section{Name Groups}
\label{sec:app-names}
To collect polysemous, rare and unknown names, we counted the number of occurrences of all-possible names in the pre-training corpus, Wikipedia\footnote{{https://huggingface.co/datasets/wikipedia}} and BookCorpus\footnote{https://huggingface.co/datasets/bookcorpus}. We denote the frequency of a name as $f_{exact}$ and $f_{ner}$ representing doing exact string match or named entity recognition when counting name occurrences respectively. Rare contains names shown at least once and with the lowest $f_{exact}$ not equaling 0. Unknown includes names with $f_{exact}$ equaling 0. According to our observations, we find that names with a larger $f_{exact}$ are likely to be polysemy and are not uniquely used as personal names. 
So, we design a metric to recognize such names as follows:
\begin{equation}
	u = \frac{rank(f_{exact})-rank(f_{ner})}{rank(f_{exact})+rank(f_{ner})}
	\label{eq:uniqueness}
\end{equation}
$rank(\cdot)$ means that the ranking of a name among the whole name list based on its frequency in descending order~\footnote{Doing named entity recognition on the whole pre-training corpus is too time-consuming. Therefore, we randomly sample 1\% of the data for counting the $f_{ner}$ and use the name rankings in Eq.~\ref{eq:uniqueness} to get the uniqueness score.}. A higher $u$ shows a higher level of uniqueness of a word as a name. The names with the lowest $u$ scores are selected as Polysemous in Sec.~\ref{sec:unfairness}.

Examples of names in different name groups are listed as follows:
\begin{itemize}
	\item \textbf{Frequent}: Alexis, Philip, Matthew, Frank, Tyler, Roy, Catherine, Joan, Amanda, Henry
	\item \textbf{Polysemous}: July, Sea, March, Paris, Treasure, Oxford, Romania, Ice, Jersey, Navy
	\item \textbf{Rare}: Makinzy, Diyanna, Javione, Zamire, Harkeem, Jerralyn, Crissi, Monque, Ajahar, Dijion
	\item \textbf{Unknown}: Jaliyiah, Cardelia, Ravindr, Josephanthony, Tyjohn, Tnaya, Jyren, Kashaunda, Jaykob, Latonnia
	\item \textbf{White}: Kim, Georgia, Joseph, Mark, Martin, James, William, Barbara, Richard, Victoria
	\item \textbf{Hispanic}: Sofia, Daisy, Luis, Manuel, Dora, Emilia, Minerva, Antonio, Oscar, Francisco
	\item \textbf{Black}: Kenya, Ebony, Anderson, Kelvin, Dexter, Cleveland, Percy, Mamie, Jarvis, Essie
	\item \textbf{Asian}: Kong, Muhammad, Gang, Mai, Chi, Krishna, Can, Wan, Wang, Ferdinand
\end{itemize}

\section{Hyper-parameter Search}
\label{sec:app-hyper}
We empirically searched the hyper-parameters $\alpha$ and $\beta$ in \{1, 10, 20\} respectively with 9 combinations for Ins. Due to the limited computation resources and the large search space, we trained the model with different combinations for a single time, selected the best 3 combinations and 
repeated experiments with different random seeds to determine the final choice of $\alpha$ and $\beta$ according to the performance on $D_{va}$. Finally, we set ($\alpha$, $\beta$) as (1, 10), (1, 10), (1,1) for dialogue summarization, question generation and reading comprehension respectively. We directly borrow these settings for FreIns.

\begin{figure}[h]
	\centering
	\includegraphics[scale=0.6]{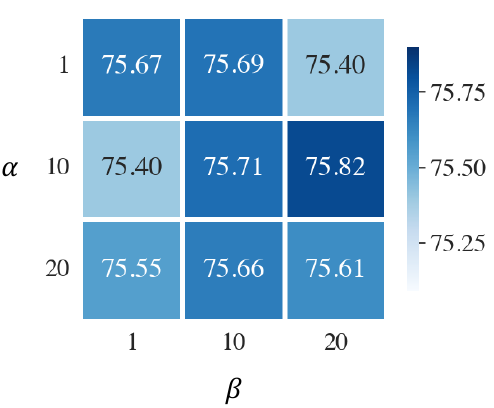}
	\caption{BertScore(\%) on the vanilla test set with different hyper-parameters.} %in the summaries.}
	%\underline{Divergent contents} are underlined in the generated summaries.}
	\label{fig:hyper}
\end{figure}

In Fig.~\ref{fig:hyper}, we show the performances of Ins under different combinations for dialogue summarization on the vanilla test set with a single run. We can see that all of the results outperform the baselines in Table~\ref{tab:mdresults-vanilla} and the standard deviation of BertScore among different combinations is only 0.14\%, showing the stable improvements of Ins over the baselines.

\begin{table}[h!]
	\scriptsize
	\centering
	\begin{subtable}{\linewidth}
		\scriptsize
		\centering
		\begin{tabular}{p{0.9cm}|p{0.36cm}p{0.36cm}p{0.36cm}p{0.36cm}|p{0.36cm}p{0.36cm}p{0.36cm}p{0.38cm}}
			\toprule[1pt]
			
			& \multicolumn{4}{c|}{\textbf{R2}} & \multicolumn{4}{c}{\textbf{BertScore}} \\
			%\cline{2-9}
			\textbf{Approach}& - & S$\downarrow$ & R$\downarrow$ & D$\downarrow$ & - & S$\downarrow$  & R$\downarrow$ & D$\downarrow$\\
			%& \multicolumn{6}{c}{\textbf{Change-one-name}} \\
			%\textbf{Approach} & \textbf{R2} & \textbf{BertS} & \textbf{D-R2} & \textbf{R-R2} & \textbf{D-BertS} & \textbf{R-BertS} \\
			
			\hline
			\multicolumn{7}{l}{\textit{In-distribution Names}}\\
			Vanilla &
			27.29 & 25.53 & 11.05& 4.42& 74.64&9.65&5.19 & 2.05\\
			Emb & 27.41 & 24.20  & 10.87 & 4.33&74.90&9.49& 5.29& 2.09  \\
			Aug&
			{27.51} &22.24   & {9.89} & {3.96}&74.83&8.50& 4.67& {1.85}  \\
			Ins$\star$  & \textbf{28.70} &\textbf{16.64}& \textbf{7.19} & \textbf{2.92} & \textbf{75.44} &\textbf{6.11} & \textbf{3.18} & \textbf{1.28}\\

			\hline
			\multicolumn{7}{l}{\textit{All-possible Names}}\\
			Vanilla &
			27.32&25.77  & 11.07 & 4.45&74.81 &9.61&5.15& 2.04  \\
			Emb & 27.26 & 24,98  & 10.68 & 4.25&74.80&9.57 & 5.16& 2.02\\
			Aug  &
			27.36 & 22.73 & {10.04} & {4.03} &74.86&8.56& 4.69& 1.87  \\
			Ins$\star$ & \textbf{28.38} &\textbf{18.65} & \textbf{8.12}& \textbf{3.29}  &\textbf{75.35} &\textbf{6.89}& \textbf{3.75} & \textbf{1.50} \\
			
			\bottomrule[1pt]
		\end{tabular}
		\caption{Dialogue Summarization}
		%\label{tab:mdresults-ds}
	\end{subtable}
	
	\begin{subtable}{\linewidth}
		\scriptsize
		\centering
		\begin{tabular}{p{0.9cm}|p{0.36cm}p{0.36cm}p{0.36cm}p{0.36cm}|p{0.36cm}p{0.36cm}p{0.36cm}p{0.36cm}}
			\toprule[1pt]
			& \multicolumn{4}{c|}{\textbf{Bleu}} & \multicolumn{4}{c}{\textbf{RL}} \\
			%\cline{2-9}
			\textbf{Approach}& - & S$\downarrow$  & R$\downarrow$ & D$\downarrow$& - & S$\downarrow$  & R$\downarrow$& D$\downarrow$ \\
			%& \multicolumn{6}{c}{\textbf{Change-one-name}} \\
			%\textbf{Approach}  & \textbf{Bleu} & \textbf{RL} & \textbf{D-Bleu} & \textbf{R-Bleu} & \textbf{D-RL} & \textbf{R-RL} \\
			
			\hline
			\multicolumn{7}{l}{\textit{In-distribution Names}}\\
			Vanilla &
			17.93& 18.76& 6.08& 2.58&{56.85} &8.17&7.55& 3.12 \\
			Emb  & 18.34 &22.22 & 7.63 & 3.26 &56.84 & 10.07 & 9.62 & 3.98\\
			Aug &
			18.06 & 14.82 & {4.39} & {1.90} &56.12 &6.91 & {6.38} & 2.69 \\
			Ins$\star$ & \textbf{19.45} &\textbf{9.66}  & \textbf{2.75} & \textbf{1.18}& \textbf{57.31}&\textbf{4.50}& \textbf{4.27} & \textbf{1.81} \\
			%Ins+ & 18.78 & 56.97 & 3.08 & 7.21 & 4.01 & 9.73 &\\
			
			\hline
			\multicolumn{7}{l}{\textit{All-possible Names}}\\
			Vanilla &
			17.91& 17.73& 5.75 & 2.46&56.67 & 7.76&7.05 & 2.95\\
			Emb  & 18.67 & 20.80  & 7.08& 3.06&56.86& 9.47& 8.89 & 3.73 \\
			Aug  &
			17.97 & 13.04 & {3.62} & {1.57} &56.12&6.06& {6.50}& {2.25}  \\
			Ins$\star$& \textbf{19.60} & \textbf{8.11} & \textbf{2.22} & \textbf{0.97}& \textbf{57.51}&\textbf{3.77}& \textbf{3.42}& \textbf{1.47} \\
			
			\bottomrule[1pt]
		\end{tabular}
		\caption{Question Generation}
		%\label{tab:mdresults-qg}
	\end{subtable}
	
	\begin{subtable}{\linewidth}
		\scriptsize
		\centering
		\begin{tabular}{p{0.9cm}|p{0.36cm}p{0.36cm}p{0.36cm}p{0.36cm}|p{0.36cm}p{0.36cm}p{0.36cm}p{0.36cm}}
			\toprule[1pt]
			& \multicolumn{4}{c|}{\textbf{Bleu}} & \multicolumn{4}{c}{\textbf{RL}} \\
			%\cline{2-9}
			\textbf{Approach}& - & S$\downarrow$  & R$\downarrow$ & D$\downarrow$& - & S$\downarrow$ & R$\downarrow$ & D$\downarrow$ \\
			%& \multicolumn{6}{c}{\textbf{Change-one-name}} \\
			%\textbf{Approach}  & \textbf{Bleu} & \textbf{RL} & \textbf{D-Bleu} & \textbf{R-Bleu} & \textbf{D-RL} & \textbf{R-RL} \\
			
			\hline
			\multicolumn{7}{l}{\textit{In-distribution Names}}\\
			Vanilla &
			27.96 &54.08& 3.85 & 1.67 & 73.91 &4.49 & 5.50 & 2.37\\
			Emb & 25.52 & 56.61 & 4.28 & 1.85&70.20& 5.32 & 6.37& 2.75 \\
			Aug  &
			26.54 &  54.76 & 3.69 & 1.60&72.53&4.57& 5.87& 2.55 \\
			Ins$\star$& \textbf{29.03} &\textbf{52.03}  & \textbf{2.48} & \textbf{1.08}& \textbf{74.81}&\textbf{5.65}& \textbf{4.41}& \textbf{1.91}  \\
			%Ins+ & 18.78 & 56.97 & 3.08 & 7.21 & 4.01 & 9.73 &\\
			
			\hline
			\multicolumn{7}{l}{\textit{All-possible Names}}\\
			Vanilla&
			27.82 &53.48 & 2.81 & 1.22 &73.97 & 3.28& 4.07& 1.77 \\
			Emb & 25.14 & 56.08 & 3.04 & 1.32 &70.51&4.31& 4.89 & 2.12 \\
			Aug &
			26.64 &  53.71 & 2.92& 1.27 &72.68&3.61& 4.61& 2.00  \\
			Ins$\star$  & \textbf{29.40} & \textbf{51.20}& \textbf{1.93} & \textbf{0.83} & \textbf{74.94}&\textbf{2.41} & \textbf{3.13}& \textbf{1.36} \\
			%Ins+ & 18.78 & 57.07 & 2.61 & 6.09 & 3.46 & 8.24 &\\
			\bottomrule[1pt]
		\end{tabular}
		\caption{Reading Comprehension}
		%\label{tab:mdresults-rc}
	\end{subtable}
	\caption{Performances(\%) of offline approaches for sensitivity on an individual speaker.}	
	\label{tab:speaker-off}
\end{table}

\begin{table}[t]
	\scriptsize
	\centering
	\begin{subtable}{\linewidth}
		\scriptsize
		\centering
		\begin{tabular}{p{0.9cm}|p{0.36cm}p{0.36cm}p{0.36cm}p{0.36cm}|p{0.36cm}p{0.36cm}p{0.36cm}p{0.38cm}}
			\toprule[1pt]
			
			& \multicolumn{4}{c|}{\textbf{R2}} & \multicolumn{4}{c}{\textbf{BertScore}} \\
			%\cline{2-9}
			\textbf{Approach}& - & S$\downarrow$ & R$\downarrow$ & D$\downarrow$ & - & S$\downarrow$  & R$\downarrow$ & D$\downarrow$\\
			%& \multicolumn{6}{c}{\textbf{Change-one-name}} \\
			%\textbf{Approach}  & \textbf{R2} & \textbf{BertS} & \textbf{D-R2} & \textbf{R-R2} & \textbf{D-BertS} & \textbf{R-BertS} \\
			
			\hline
			%Vanilla & 28.12 & 75.09 & - &-&-&-&-&-&-&-&-&-\\
			%ID &-&-&-&-&-&-&-&\\
			Fre  &
			{28.40} &20.28 & 9.25 & 3.73 &75.10 &7.85& 4.29& 1.72  \\
			FreAug  &
			27.91 &  20.11 & {9.02}& {3.64}&{74.97}&7.78& 4.24 & 1.70 \\
			FreIns$\star$ & \textbf{28.58} &\textbf{13.29 }  & \textbf{5.99}& \textbf{2.46} &\textbf{75.42}&\textbf{4.91} & \textbf{2.68} & \textbf{1.09}\\
			\bottomrule[1pt]
		\end{tabular}
		\caption{Dialogue Summarization}
		%\label{tab:ddresults-ds}
	\end{subtable}
	
	\begin{subtable}{\linewidth}
		\scriptsize
		\centering
		\begin{tabular}{p{0.9cm}|p{0.36cm}p{0.36cm}p{0.36cm}p{0.36cm}|p{0.36cm}p{0.36cm}p{0.36cm}p{0.36cm}}
			\toprule[1pt]
			& \multicolumn{4}{c|}{\textbf{Bleu}} & \multicolumn{4}{c}{\textbf{RL}} \\
			%	\cline{2-9}
			\textbf{Approach}& - & S$\downarrow$ & R$\downarrow$ & D$\downarrow$ & - & S$\downarrow$  & R$\downarrow$& D$\downarrow$ \\
			%& \multicolumn{6}{c|}{\textbf{Change-all-name}} & \multicolumn{6}{c}{\textbf{Change-one-name}} \\
			%\textbf{Approach} & \textbf{Bleu} & \textbf{RL} & \textbf{D-Bleu} & \textbf{R-Bleu} & \textbf{D-RL} & \textbf{R-RL} \\
			
			\hline
			%Vanilla & 18.57 & 56.40 & - &-&-&-&-&-&-&-&-&-\\
			%ID &-&-&-&-&-&-\\
			Fre &
			{18.90} &10.59& 2.97 & 1.29 & 57.01 & 4.76& 4.09 & 1.74\\
			FreAug  &
			18.60 & 8.62 & 2.54 & 1.10 &{57.13}&3.81& 3.46& 1.48  \\
			FreIns$\star$  & \textbf{19.29} & \textbf{5.48}  & \textbf{1.76}& \textbf{0.77} &\textbf{56.91}&\textbf{2.39} & \textbf{2.18}& \textbf{0.94}\\
			\bottomrule[1pt]
		\end{tabular}
		\caption{Question Generation}
		%\label{tab:ddresults-qg}
	\end{subtable}
	
	\begin{subtable}{\linewidth}
		\scriptsize
		\centering
		\begin{tabular}{p{0.9cm}|p{0.36cm}p{0.36cm}p{0.36cm}p{0.36cm}|p{0.36cm}p{0.36cm}p{0.36cm}p{0.36cm}}
			\toprule[1pt]
			& \multicolumn{4}{c|}{\textbf{Bleu}} & \multicolumn{4}{c}{\textbf{RL}} \\
			%\cline{2-9}
			\textbf{Approach}& - & S$\downarrow$ & R$\downarrow$ & D$\downarrow$ & - & S$\downarrow$ & R$\downarrow$ & D$\downarrow$ \\
			%& \multicolumn{6}{c|}{\textbf{Change-all-name}} & \multicolumn{6}{c}{\textbf{Change-one-name}} \\
			%\textbf{Approach} & \textbf{Bleu} & \textbf{RL} & \textbf{D-Bleu} & \textbf{R-Bleu} & \textbf{D-RL} & \textbf{R-RL} \\
			
			\hline
			%Vanilla & 28.42 & 73.33 & - &-&-&-&-&-&-&-&-&-\\
			%ID &-&-&-&-&-&-\\
			Fre &
			27.15 & 53.86 & 2.07 & 0.89 &73.89&2.67& 3.22 & 1.39 \\
			FreAug  & 27.82 & \textbf{52.03}& 1.83 & 0.80 &74.32 & 2.33& 3.08 & 1.33\\
			FreIns$\star$ & \textbf{28.57} & 52.41  & \textbf{1.46}& \textbf{0.64} &\textbf{74.89}&\textbf{1.70}& \textbf{2.36}& \textbf{1.02}  \\
			
			\bottomrule[1pt]
			
		\end{tabular}
		\caption{Reading Comprehension}
		%\label{tab:ddresults-rc}
	\end{subtable}

	\caption{Performances(\%) of online approaches for sensitivity on an individual speaker.}	
	\label{tab:speaker-on}
\end{table}

\begin{figure}[h!]
	\centering
	\includegraphics[width=0.95\linewidth]{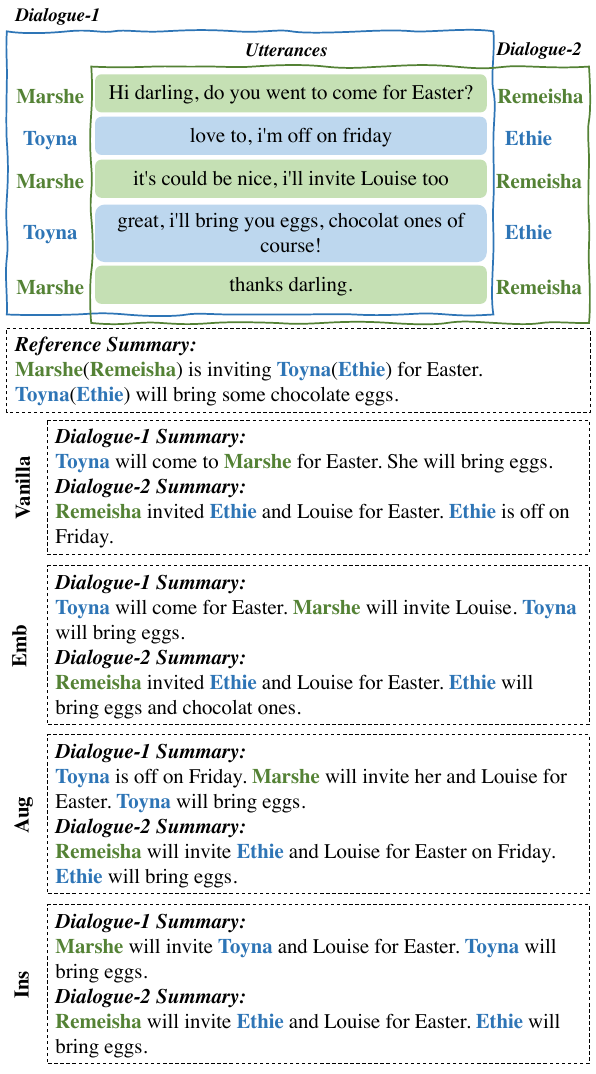}
	\caption{Case study for dialogue summarization.} 
	\label{fig:case-ds}
\end{figure}

\section{Additional Results of Sensitivity on an Individual Speaker}
\label{sec:app-results}
Results for sensitivity on an individual speaker on all of the three tasks are in Table~\ref{tab:speaker-off} and Table~\ref{tab:speaker-on}. Both tables lead to the same observations and conclusions as discussed in Sec~\ref{sec:mda} and Sec~\ref{sec:onlineapproach}, where Ins and FreIns perform best among offline and online approaches correspondingly.

\section{Case study}
\label{app:casestudy}

We show cases for different tasks in this section.

\begin{figure}[t!]
	\centering
	\includegraphics[width=0.95\linewidth]{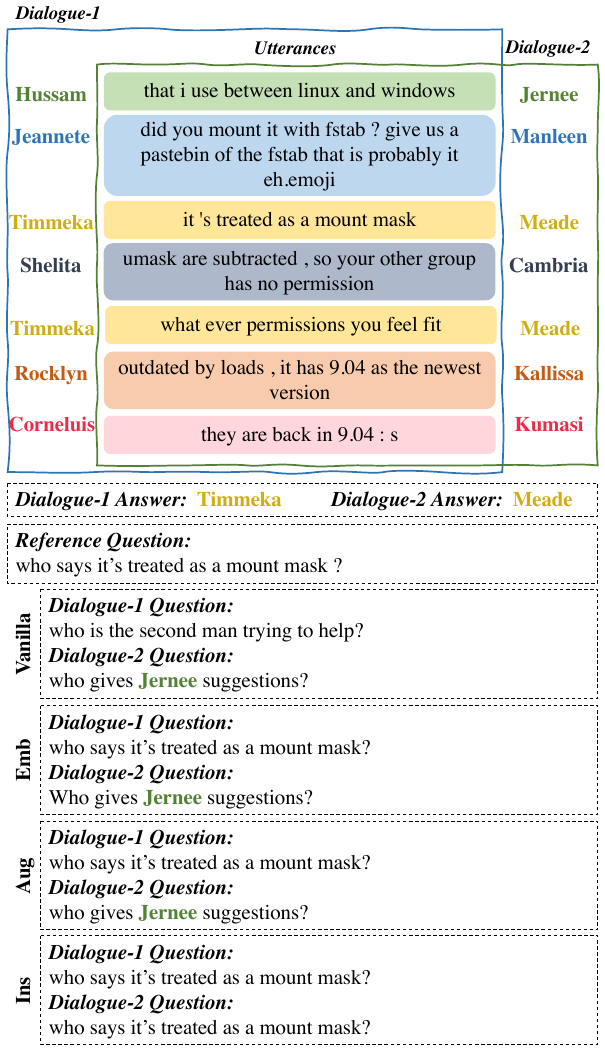}
	\caption{Case study for question generation.} 
	\label{fig:case-qg}
\end{figure}

\begin{figure}[h]
	\centering
	\includegraphics[width=0.95\linewidth]{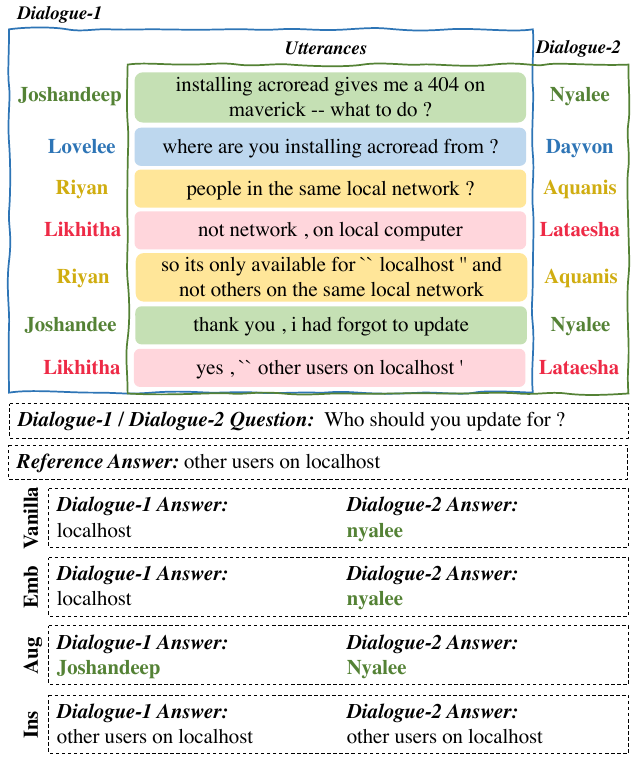}
	\caption{Case study for reading comprehension.} 
	\label{fig:case-rc}
\end{figure}

The case for dialogue summarization is in Fig.~\ref{fig:case-ds}. Vanilla extracts different information for two sets of names: ``She will bring eggs'' and ``Ethie is off on Friday''. It also uses different expressions: ``will come to ... for Easter'' and ``invited ... for Easter''. Besides, ``Louise'' is only mentioned in the second summary.
Emb has the information difference and the expression difference. Meanwhile, it outputs incorrect content in the second summary, where ``chocolat ones'' is used for describing ``eggs'' in the input dialogue. 
Aug outputs more information for the first set of names. Ins treats the two sets of names equally with the same generations modulo the speaker names.

In the case of question generation in Fig.~\ref{fig:case-qg}, all baselines generate ``who gives Jernee suggestions?'' for the second set of names, which is an inaccurate question with multiple candidate answers. Emb also generates a ``Who'' with the capitalized first letter, which is also different from the other one with lowercase ``who'' if we compare them strictly. Ins generates identical and accurate questions for the same dialogue with different speaker names.

For reading comprehension in Fig.~\ref{fig:case-rc}, both Vanilla and Emb generate quite different answers for two sets of names. Aug generates consistent but wrong answers considering the one-to-one mapping of speaker names. Ins outputs identical correct and complete answers, outperforming the baselines.

\end{document}